\documentclass{article}

\PassOptionsToPackage{numbers, compress}{natbib}
\usepackage[authoryear, sort&compress, round]{natbib}
\usepackage{booktabs}
\usepackage{tabularx}
\usepackage{array}
\usepackage{graphicx}
\usepackage[utf8]{inputenc}
\usepackage{booktabs}
\usepackage{enumitem}
\usepackage{amssymb}
\usepackage{titlesec}
\newcommand{\checkbox}{$\square$}
\newcommand{\radio}{$\bigcirc$}

\usepackage[creativeai, final]{neurips_2026}

\usepackage[utf8]{inputenc} %
\usepackage[T1]{fontenc}    %
\usepackage{hyperref}       %
\usepackage{url}            %
\usepackage{booktabs}       %
\usepackage{amsfonts}       %
\usepackage{amsmath}        %
\usepackage{nicefrac}       %
\usepackage{microtype}      %
\usepackage[table]{xcolor}  %
\usepackage{subcaption}     %

\newcount\Comments  %
\usepackage{color}
\definecolor{darkred}{rgb}{0.9,0,0}
\definecolor{darkgreen}{rgb}{0,0.5,0}
\definecolor{darkblue}{rgb}{0,0,0.7}
\definecolor{purple}{rgb}{.6, 0,.6}
\definecolor{orange}{rgb}{1.0,0.64,0}
\newcommand{\kibitz}[2]{\ifnum\Comments=1\textcolor{#1}{#2}\fi}

\title{Shaping AI’s Beliefs Before It Generates: \\Human Sense of Agency in Creative AI}

\author{%
  Zi Wang\thanks{The Proactive Co-Creator (PaCC) can be accessed at \url{https://proactive-co-creator.github.io/}. } \\ %
    wangzi@google.com \\
  Google DeepMind \\
  \And
  Mahima Pushkarna \\
mahimap@google.com \\
  Google DeepMind \\
}

\begin{document}

\maketitle

\begin{abstract}
AI systems for creative media generation expand human capabilities yet risk diminishing human agency. A core problem is anchoring, where the user's thinking becomes anchored to a polished, generated artifact, shifting the user's role from active authoring to reactive editing. We propose \emph{pre-generation belief controls} that can empower creators to inspect, revise, and steer the AI's interpretation \emph{before} any media is generated, so that the initial anchor is set by the human. We formalize four design principles and instantiate them in the Proactive Co-Creator (PaCC), a system for early-stage conceptual ideation and prototyping that proactively surfaces its assumptions and uncertainty through clarifications, belief graphs, and media attributes, inviting users to shape the AI's beliefs before generating the final outcome. Using PaCC as a design probe in a small-scale, single-arm exploratory user study, we observed that participants reported directionally positive sense of agency and process control relative to their familiar prompt-based AI tools, noting that surfaced assumptions helped them think through details they had overlooked, whereas their feelings about ownership and authorship diverged.%
\end{abstract}
\begin{figure}
    \centering
    \includegraphics[width=1\linewidth]{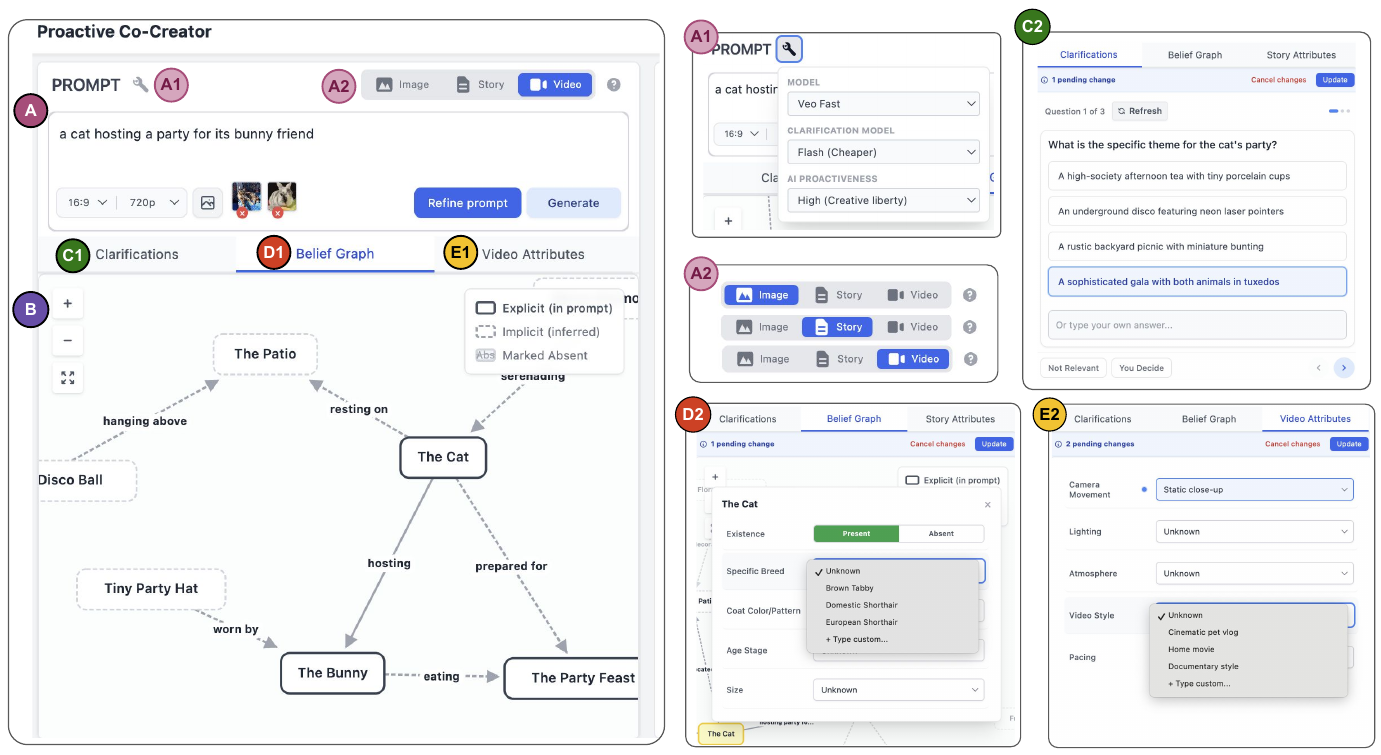}
    \caption{The Proactive Co-Creator (PaCC) interface for shaping AI's beliefs before generation (\S\ref{sec:pacc}). (A)~Prompt entry with optional reference images. Creators control AI proactiveness (A1) across three tiers: literal extraction (\emph{low}), reasonable scene inference (\emph{medium}; e.g., adding a ``party table'' to the cat party), or creative liberty (\emph{high}; e.g., a ``disco ball''). Modality tabs switch between Image, Story, and Video (A2). (B)~Belief Graph: An editable visualization of explicit entities (solid borders), AI-inferred entities (dashed borders), and relationship edges (e.g., ``hosting''). (C1--C2)~Clarification Questions: Offers 3--5 selectable options per question so creators can resolve ambiguity and guide the AI's interpretation; clicking ``You Decide'' delegates a specific choice to the AI. (D1--D2)~Entity Attribute Editing: Adjusts properties of individual nodes. (E1--E2)~Media Attributes: Controls global scene aesthetics (e.g., lighting, style, pacing). By surfacing and aligning assumptions and possibilities prior to media generation, PaCC establishes the human intent as the anchor.}
    \label{fig:pacc_illustration}
\end{figure}

\section{Introduction}
\label{sec:introduction}
Human agency, the capacity to make deliberate choices and control one's actions~\citep{sep-agency}, is central to well-being~\citep{hojman2018agency, welzel2010agency, haggard2017sense} and paramount in creative work to ensure decisions reflect authentic vision and intent. Generative AI expands human capabilities~\citep{united2025human}, yet risks diminishing human sense of agency (SoA). Typical human-AI interactions rely on \emph{post-generation critique}, where users prompt a model, receive a polished artifact, and then iteratively edit (or re-prompt) the system, positioning users as reactive curators rather than active authors. 
These workflows can induce an anchoring effect on users~\citep{furnham2011literature}. By autonomously resolving unspecified details and making choices, models can bias human reasoning so that subsequent iterations orbit the AI's initial generation rather than exploring authentic intent~\citep{shaw2026thinking, fernandes2025ai, storey2026technical}. This can contribute to reduced ownership~\citep{he2026subjective}, destabilized SoA~\citep{rafner2025agency}, and homogenized creative identity~\citep{jiang2026artificial}, detailed in \S\ref{sec:background}.

We posit that this erosion of agency stems in part from \emph{when} intervention occurs. In contrast to post-generation critique, we propose \emph{pre-generation belief control}: empowering creators to inspect, revise, and steer the AI's underlying conceptual interpretation of human intent \emph{before} media is generated, so that the initial anchor is largely determined by the human rather than the machine.

We formalize this approach into four design principles adapted from~\citet{pair_guidebook}: intervening before generation (DP1), making assumptions visible (DP2), preserving the possibility space (DP3), and calibrating AI autonomy (DP4). We instantiate these principles in the \emph{Proactive Co-Creator} (PaCC; Figure~\ref{fig:pacc_illustration}), an interactive multi-modal system for early-stage conceptual ideation and prototyping across text, image, and video. Rather than treating prompt interpretation as an opaque one-shot step, PaCC adopts three pre-generation mechanisms: (i)~proactive clarification dialogues, (ii)~editable belief graphs that distinguish user-specified from AI-inferred entities and show their relationships, and (iii)~scene-level media attributes that expose global stylistic defaults.

To explore how creators interact with these mechanisms, we deployed PaCC as a design probe in a small-scale, single-arm exploratory user study, where observations are preliminary and hypothesis-generating rather than confirmatory. While participants felt PaCC's pre-generation controls surfaced hidden assumptions and supported agency and process control, their views on ownership and authorship diverged based on their initial creative intent, stance on AI training data, and text-only style limitations. These observations highlight key design tensions and motivate testable hypotheses for future work on controlled evaluations, multimodal style steering, and agency-aware training.

In summary, our contributions are:
(1)~the concept of \emph{pre-generation belief control} and four design principles for supporting human agency in early-stage ideation;
(2)~the \emph{Proactive Co-Creator} (PaCC) system, instantiating these principles via clarification dialogues, editable belief graphs, and scene-level attributes; and
(3)~exploratory insights from a design probe that surface key design tensions and outline future research directions for agency, ownership, and authorship in human-AI co-creation.

\section{The Proactive Co-Creator}
\label{sec:pacc}

The Proactive Co-Creator (PaCC) instantiates our design principles (\S\ref{sec:introduction}) as an interactive, multi-modal system for image, story, and video generation, extending designs in our prior work~\citep{hahn2024proactive}. PaCC is a React/TypeScript web application, built with Gemini assistance in Google AI Studio, that orchestrates structured Gemini API calls for belief graph construction, clarification, and prompt refinement (Appendix~\ref{ssec:compute}). %
At its core, PaCC inserts an interactive belief control phase between prompt entry and media generation (\textbf{DP1: Intervene before generation}), shifting human-AI iteration from post-hoc editing to upstream conceptual guidance.

Figure~\ref{fig:pacc_illustration} illustrates the PaCC interface. To support human agency across creative intent, expression, and process, PaCC operationalizes our design principles through three core pre-generation interventions:
\vspace{-1em}
\paragraph{Clarification dialogues and delegation (Figure~\ref{fig:pacc_illustration}~C1--C2).}
PaCC surfaces targeted, single-topic clarifying questions to resolve scene ambiguities (Figure~\ref{fig:pacc_illustration}~C1--C2). Surfacing 3--5 diverse candidate options per question \textbf{preserves the creative possibility space (DP3)}, encouraging exploration rather than collapsing decisions into singular AI defaults. To \textbf{calibrate AI autonomy (DP4)}, each question includes a ``You Decide'' response (letting the AI decide), allowing creators to delegate specific tactical choices (e.g., color scheme) to the model while retaining manual control over core creative decisions (e.g., mood and setting).
\vspace{-1em}
\paragraph{Belief graphs, relationship editing, and entity attributes (Figure~\ref{fig:pacc_illustration}~B, D1--D2).} 
To \textbf{make assumptions visible (DP2)}, PaCC parses the user's prompt (including optional reference images) into an editable graph distinguishing \emph{explicit entities} directly stated by the user (solid borders; e.g., ``cat,'' ``bunny'') from \emph{implicit entities} inferred by the model (dashed borders; e.g., ``patio''), alongside editable relationships (edge values; e.g., ``hosting''). Each entity node carries fine-grained attributes with multiple candidate values (Figure~\ref{fig:pacc_illustration}~D1--D2; e.g., cat breed, party theme), \textbf{preserving the creative possibility space (DP3)} further inviting exploration. While the backend estimates model uncertainty for those candidate values, raw numeric probabilities are omitted from the UI to prevent cognitive clutter. Additionally, creators can \textbf{calibrate AI autonomy (DP4)} over these graph inferences via three proactiveness tiers (\emph{low}, \emph{medium}, \emph{high}; Figure~\ref{fig:pacc_illustration}~A1), which govern how aggressively the model populates implicit entities, relationships, and candidate attributes.
\vspace{-1em}
\paragraph{Media attributes (Figure~\ref{fig:pacc_illustration}~E1--E2).}
Beyond individual entities, PaCC surfaces macro, modality-specific media attributes, such as lighting, art style, and camera angle for images; narrative tone for stories; and camera motion, pacing, and duration for videos (Figure~\ref{fig:pacc_illustration}~E1--E2). Making these aesthetic defaults visible (\textbf{DP2}) and providing alternative stylistic choices \textbf{preserves the creative possibility space (DP3)}, preventing hidden style biases. As with the belief graph, the proactiveness tier setting dynamically modulates these aesthetic inferences, enabling creators to \textbf{calibrate AI autonomy (DP4)} over global stylistic defaults.

\vspace{0.3em}
Before triggering media generation (\textbf{DP1}), PaCC compiles all graph edits, clarification answers, and attribute selections into an AI-refined prompt diff for optional verification. Creators can use these pre-generation controls across iterative workflows spanning Image, Story, and Video modes. The full technical implementation details, including LLM call pipelines, incremental JSON schemas, and a step-by-step user journey walk-through, are provided in Appendix~\ref{ssec:compute} alongside a principle-to-feature mapping (Table~\ref{tab:principle_feature_map}).

\section{Measuring Human Sense of Agency in AI-assisted Creative Practice}
\label{sec:measuring_agency}

To evaluate whether PaCC's pre-generation belief controls affect human agency, we measure a \emph{subjective} sense of agency (SoA) rather than objective metrics, because an individual's felt agency can neither reliably track individual contributions nor be calculated from the division of labor between user and AI~\citep{limerick2014experience, synofzik2008beyond, rafner2025agency, he2026subjective}. Drawing on the SoA literature~\citep{moore2016sense, gallagher2012multiple, polito2013developing}, we define SoA as ``the feeling of being in control of one's own actions and their outcomes,'' and assess whether creators recognize the final artifact as their own creations~\citep{haggard2017sense, moore2016sense}.

We examine SoA alongside three interrelated experiential dimensions relevant to the context of our study\footnote{While there are several other components that contribute to artistic sense of agency identified in literature, we consider these beyond the scope of this work.}: \emph{Control} or the felt ability to direct the unfolding creative process, including exploration, intervention, and iteration. Consistent with prior work~\citep{louie2020novice, pierce2003state}, we treat control as a distinct measure from ownership, though a sense of control contributes to ownership. \emph{Ownership}, or the feeling that a work,  including process and outcome, is one's own and closely connected to the self~\citep{pierce2003state}. This is considered distinct from SoA because creators may feel an artifact reflects their taste and sits within their body of work, without feeling that they produced it~\citep{gallagher2000philosophical}. \emph{Authorship}, or the willingness to claim public credit for the work.

Because automated tooling actively influences creative workflows~\citep{heigl2025generative}, creators need to continuously negotiate and align AI contributions with their evaluative criteria and artistic values. Extending~\citet{rhodes1961analysis}, we evaluate these agency dimensions across four general aspects of creative work: \textbf{Intent} (brainstorming and forming concepts for creative work), \textbf{Expression} (exploring or refining unique artistic techniques, stylistic choices, voice or tone), \textbf{Process} (the non-linear creative process of switching between exploration and production~\citep{botella2018stages}), and \textbf{Outcome} (the media artifacts produced during and as a final result of the process~\citep{epstein2020gets}). Theoretical definitions and evaluation criteria for these are detailed in Appendix~\ref{sec:appendix_creative_aspects}.

\section{Exploratory User Study and Results}

We deployed PaCC as a design probe to examine how creators ($N{=}8$ participants, 2 female, 6 male) interact with pre-generation belief controls and to generate hypotheses for future controlled studies.    Participants (Table~\ref{tab:participants}) were recruited from within our institution and had used generative AI for creative ideation or artistic exploration within the past 6 months. 5 participants worked in AI research and engineering (P1--P3, P6, P8), 2 were practicing creative interaction designers (P5, P7), and 1 in UX research (P4). Following a 10-minute demo, participants spent 30 minutes on a self-directed multi-modal generation task in PaCC across text, image, and video. They then completed a 20-minute questionnaire rating PaCC overall and its features across sense of agency, control, ownership, and authorship on a 5-point Likert scale ($1{=}\text{diminished}$, $3{=}\text{preserved/no change}$, $5{=}\text{augmented}$), alongside retrospective comparisons against their go-to prompt-based AI tools (Appendices~\ref{sec:appendix_study_details}--\ref{sec:user_study_form}). Participants were provided with a broad definition of SoA for ratings. A semi-structured discussion followed the survey.

\label{ssec:user_study_design}

Given the small sample ($N{=}8$) and single-arm design, we treat all findings as exploratory and directional rather than statistically conclusive. We focus on qualitative insights from participant feedback, reporting quantitative Likert ratings as $\text{mean} \pm \text{standard deviation}$ unless otherwise specified.

\vspace{-.5em}
\paragraph{Defining sense of agency for AI co-creation.} Participants most frequently defined SoA in terms of ``process control'' (6), ``decision control'' (4), and ``ownership'' (3), as well as ``\textit{the ability to influence the final output}'' (P1), spanning three recurring themes:

\begin{itemize}[leftmargin=*, itemsep=0pt, topsep=1pt]
    \item \textbf{Control and Ownership}. Being in charge of the creative process, initiating actions, and maintaining ownership of the final output even when collaborating with AI: \textit{"Feels like I'm in charge of what I'm building out especially with AI tools - where I am in command and AI is collaborating with me not taking my sense of ownership away. It's still directed and owned by me. I feel like I'm part of the creative process."} --P8
    \item \textbf{Intent and Execution}. The ability to effectively communicate intent, have the AI reliably follow instructions, and steer the creative process through iterations: \textit{"In the context of GenAI my sense of agency would relate to how effectively I can communicate my intent to the AI and have it reliably follow my instructions."} --P2
    \item \textbf{Empowerment and Expression}. Having the freedom to express thoughts without friction and possessing the power to make informed choices during creative work: \textit{"Freedom to express my thoughts without friction and feeling heard. That being said not every creative thought is about re-imagining the world, most of my creative thinking is quite tactical. I've already felt the problem and have an idea what I want to make so I can iterate on it."} --P7
\end{itemize}
\vspace{-0.5em}
\paragraph{AI alignment and past experience can influence SoA.} Participants described their current AI tools as having an overall neutral effect on their SoA ($\text{diminish}{=}3.13 \pm0.35$, $\text{preserve}{=}3.13\pm0.64$, $\text{augment}{=}3.00\pm0.76$), suggesting that current AI tools neither consistently preserve nor undermine users' SoA. %
Open-ended responses noted that SoA diminished when tools felt invasive or unaligned: \textit{"If it's not hijacking my sense of creativity, AI tools do augment me in the right way if they don't feel too invasive; they can diminish if they don't understand what I want after multiple attempts"} --P8, or when outputs had evidence of the training data, \textit{"I fundamentally don't feel any ownership over any AI outputs, especially like in one of my explorations ... the [anonymized] logo appeared in the output just as a constant reminder of the training data from real creatives used in the models."} --P6.

\vspace{-1em}
\paragraph{More ways to control AI can enhance SoA.} Participants described how the modular and interactive nature of PaCC expanded creative possibilities by offering a wider variety of choices, \textit{"I liked being able to edit the belief graph / modify my prompt with clarifying questions. The modular nature was also nice."} --P1, \textit{"I was able to have more agency than usual because I had more options, and the system felt more open to my input instead of doing its own thing regardless of what I say."} --P3. One designer noted that surfaced assumptions illuminated blind spots: \textit{"The clarifying steps illustrated for me the things I didn't think through, highlighted my assumptions"} --P7. 
Descriptively, participants rated PaCC positively on overall SoA ($4.38 \pm 1.06$), with mean ratings for all features above the ``3. Preserved'' midpoint ($\text{Prompt Refinement}{=}4.25 \pm 0.89$, $\text{Clarifying Questions}{=}4.00 \pm 0.93$, $\text{Media Attributes}{=}3.88 \pm 0.83$, and $\text{Belief Graph}{=}3.75 \pm 1.04$\footnote{P7 reported not using the belief graph or media attributes feature during the study.}).
\vspace{-1em}
\paragraph{Diverging views on creative ownership and authorship.} PaCC was rated most positively for intent ($3.67$), followed by process ($3.57$) and outcome ($3.50$). Creative \emph{process} received the highest SoA ratings ($4.25\pm0.89$). 
Participants highlighted the difficulties in mapping visual style using language alone: ``\emph{It was harder to map styles from my head into the photo, but relational concepts were great}'' --P1, ``\emph{Without providing an input of my own work, only using language, I feel like I'm yielding most creative styling choices to the model.}'' --P5. While intent and outcome ratings were identical for SoA ($4.12$), ownership was slightly higher for intent than for outcome ($3.50$ vs.\ $3.12$), suggesting that some participants felt greater ownership over their ideas than over the final artifact. \textit{"I started to think more about the process of creation - what am I missing, is this prompt right, the product notifications took me out of creative mindset - than the artifact. It inserted itself into my creativity. Maybe this is good, maybe this is bad. I don't know."} --P7. 
Views on ownership and authorship were polarized across participants. Those who felt little-to-no ownership over the final artifact attributed this either to open-ended exploration without a pre-formed concept or to feeling steered by AI-generated multiple-choice options: \textit{"...as far as ownership, like, do I feel like I made this? Definitely not, but especially um I think I might have felt more ownership if I had the idea fully myself and we're using the tool to try to arrive at it as opposed to exploring."} --P5, \textit{"...the final product, I definitely felt no ownership over because like at every step of the way, I was like prompted with some options I can choose between. But then by the end I like don't know like if this was what I wanted in the first place."} --P4. In contrast, half of the participants (P1, P2, P3, P8) reported high felt ownership ($M \in [4.00, 5.00]$) and preserved or augmented authorship (P1 did not rate authorship): \textit{"(Feeling of SoA, Ownership, Control) seem similar, perhaps with the exception of authorship, which feels more like it doesn't change because \textbf{I am still the author regardless}."} --P3, \textit{"I felt ownership over the end-product and the idea, but there were some style elements where I feel like reference images could have helped me confirm that I should update a particular attribute."} --P1.
Overall, PaCC received positive mean ratings for \emph{sense of agency} ($4.09\pm0.84$) and \emph{sense of control} ($3.89\pm0.69$, $N{=}7$) across intent, expression, process, and outcome, whereas ratings for \emph{sense of ownership} ($3.22\pm1.49$) and \emph{sense of authorship} ($2.75\pm1.13$, $N{=}7$) exhibited the widest variance across individuals. Together with qualitative feedback, this suggests that participants generally perceived pre-generation controls as supporting agency and process control, whereas felt ownership and authorship varied with participants' task goals and attitudes toward AI.

\vspace{-1em}
\paragraph{Pre-generation controls with interleaved interactions may enhance felt authorship.} Although the \emph{Belief Graph} earned high SoA ratings ($3.75\pm0.95$; Table~\ref{tab:combined_features}), opinions were split between it being ``useful'': ``\emph{It gives me confidence that the complex visual concepts in the scene are exactly how I picture them}'' --P1, and confusing: ``\emph{I love data visualization, but it was difficult to interact with and the structure of it didn't reflect how I was thinking}'' --P7. \emph{Prompt Refinement} scored highest across agency and control, whereas \emph{Attributes} received moderate scores partly from lower interaction. \emph{Clarifying Questions} rated uniformly high across agency, control, and ownership, and was the \textit{only} feature all 8 participants rated as providing greater agency than their current tools (Table~\ref{tab:combined_features}). The majority of participants rated PaCC above their current tools for sense of agency ($5${/}$7$), control ($6${/}$8$), and ownership ($4${/}$7$) (Table~\ref{tab:combined_features}). \emph{Authorship} was the most varied dimension ($3${/}$7$ greater, $3${/}$7$ equal, $1${/}$7$ lesser). Interestingly, individual features showed a more differentiated pattern: clarifying questions and prompt review individually received ($5${/}$7$) ``greater'' authorship ratings, while the belief graph and media attributes were more evenly split. This variation suggests that pre-generation interventions that involve active dialogue may support felt authorship more effectively than others, though this proposition would require verification with larger samples.

\begin{table}[htbp]
\centering

\caption{Mean Likert ratings ($M \pm \text{SD}$ on a 1--5 scale; $3{=}\text{preserved}$) and the number of participants rating each PaCC feature as providing \textbf{G}reater/\textbf{E}qual/\textbf{L}esser (G/E/L) agency, ownership, control, or authorship than their current AI tools. $N$ for G/E/L varies because some participants selected ``N/A.''}
\label{tab:combined_features}
\scriptsize
\begin{tabular}{@{} l cc cc cc cc @{}}
\toprule
& \multicolumn{2}{c}{Agency} & \multicolumn{2}{c}{Ownership} & \multicolumn{2}{c}{Control} & \multicolumn{2}{c}{Authorship} \\
\cmidrule(lr){2-3} \cmidrule(lr){4-5} \cmidrule(lr){6-7} \cmidrule(lr){8-9}
Feature & $M \pm SD$ & G/E/L & $M \pm SD$ & G/E/L & $M \pm SD$ & G/E/L & $M \pm SD$ & G/E/L \\
\midrule
PaCC (overall)       & $4.09 \pm 0.84$ & 5/2/0 & $3.22 \pm 1.49$ & 4/2/1 & $3.89 \pm 0.69$ & 6/2/0 & $2.75 \pm 1.13$ & 3/3/1 \\
Clarifying Questions & $3.66 \pm 0.64$ & 8/0/0 & $3.34 \pm 0.77$ & 6/0/1 & $3.66 \pm 0.79$ & 7/1/0 & $3.18 \pm 0.72$ & 5/1/1 \\
Belief Graph         & $3.75 \pm 0.95$ & 4/0/2 & $3.46 \pm 0.80$ & 4/1/2 & $3.54 \pm 0.98$ & 5/1/1 & $3.08 \pm 0.79$ & 3/2/2 \\
Attributes           & $3.29 \pm 0.81$ & 4/2/0 & $3.29 \pm 0.81$ & 3/3/1 & $3.43 \pm 0.61$ & 4/3/0 & $3.04 \pm 0.64$ & 3/3/1 \\
Prompt Refinement    & $3.84 \pm 1.04$ & 6/1/0 & $3.53 \pm 1.21$ & 6/0/1 & $3.72 \pm 1.32$ & 6/2/0 & $3.32 \pm 1.37$ & 5/1/1 \\
\bottomrule
\end{tabular}
\end{table}

\section{Discussion}
\label{sec:discussion}
We discuss the implications of our empirical observations, nuances around agency and authorship, hypotheses for future investigation, and study limitations (with extended analyses in Appendix~\ref{sec:appendix_extended_discussion}).
\paragraph{Proactive artificial agency stemming from pre-generation belief control.}
Rather than autonomously resolving unspecified details, PaCC is designed for proactive self-transparency: it identifies ambiguity, surfaces assumptions for critique, and invites creators to reshape beliefs before media is generated. This mirrors professional workflows where intermediate low-fidelity artifacts elicit critique without prematurely anchoring on a final output~\citep{dow2010parallel}. Artificial agency is thus exercised by \emph{asking} rather than \emph{answering}, allocating autonomy to process-level structure while leaving content-level creative decisions to users.
\vspace{-1em}
\paragraph{Diverging perspectives on ownership and authorship.}
While participants generally reported positive agency and process control with PaCC, their views on ownership and authorship diverged. Half of our participants reported high ownership, whereas the other half reported low ownership due to open-ended exploration without a pre-formed concept, feeling steered by AI multiple-choice options, skepticism toward AI training data provenance, or text-only style limitations. This motivates testable hypotheses for future work: (1)~pre-generation belief controls may yield stronger ownership in goal-directed tasks than in open-ended exploration, and (2)~pairing conceptual controls with multimodal style inputs (e.g., personal sketches or style references) may strengthen felt authorship over outcomes. Moreover, because aggregate agency scores can mask polarization in ownership and authorship, evaluations should measure these dimensions separately across creative aspects.
\vspace{-1em}
\paragraph{Incentivizing human agency in AI training and agent objectives.}
Models optimized solely for output quality and preference alignment can generate polished artifacts without improving SoA. To preserve human agency, future reward models and multi-turn training should incentivize collaborative negotiation, such as asking clarifying questions and surfacing implicit assumptions before generation.
\vspace{-1em}
\paragraph{Study limitations and future work.}
While our sample ($N{=}8$, primarily technical professionals) served as an initial probe into conceptual framing, longitudinal deployments with professional artists and writers are needed to examine sustained workflows, creative identity, and domain-specific agency thresholds. Furthermore, whereas our evaluation used retrospective self-assessments against familiar tools, future work should conduct controlled A/B experiments against post-generation critique baselines (further details in Appendix~\ref{sec:appendix_limitations}).

\section{Conclusion}

We proposed \emph{pre-generation belief control} to mitigate \emph{anchoring} in creative workflows, instantiated in the Proactive Co-Creator (PaCC), an interactive system for inspecting, clarifying, and steering the AI's assumptions before generating media. By surfacing model beliefs through clarification dialogues, belief graphs, and media attributes, PaCC directs artificial agency toward eliciting user intent rather than autonomously resolving unspecified details. Using PaCC as a design probe in an exploratory study, we observed that participants generally reported positive perceived agency and process control while expressing diverging perspectives on ownership and authorship, motivating testable hypotheses for future research. Ultimately, we argue that designing generative AI around the human creative process, rather than single-turn output generation alone, is important for supporting human agency.

\begin{ack}
We thank the Google AI Studio team for quality assurance testing. We would also like to thank Rich Galt, James Wexler, Been Kim, Yizheng Huang, Neha Kalibhat, Mani Malek, Noah Fiedel, Fernanda Viegas, Martin Wattenberg and many others for support, discussions, review and/or feedback.
\end{ack}

\bibliography{refs}

\appendix

\section{Background}
\label{sec:background}

\paragraph{Agency and Sense of Agency.}

In the philosophy of action, \emph{agency} denotes the capacity of an individual to make deliberate choices and exercise control over their actions~\citep{sep-agency, anscombe1956intention, davidson1963actions}. In sociology, anthropology, and cognitive science, agency is recognized not as a static property or the unilateral imposition of a pre-formed mental plan onto passive matter, but as a dynamic process of engagement with one's environment~\citep{alexander1988action, emirbayer1998agency, barandiaran2009defining}. Theories of craft and art-making emphasize that creativity emerges through an ongoing, dialogic \emph{correspondence} between creator, tool, and material~\citep{sennett2008craftsman, ingold2013making}. In human--computer interaction (HCI) and creativity support tools (CSTs), researchers examine how interface design distributes power and control between human and system~\citep{li2023beyond}, focusing on the subjective experience of agency, termed \emph{Sense of Agency} (SoA)---the feeling of being in control of one's actions and their resulting outcomes~\citep{moore2016sense, friston2013anatomy, david2015sense, polito2013developing}. \citet{gallagher2012multiple} noted that SoA involves consciously attributing an action to oneself and experiencing self-authorship when outcomes align with personal intentions, beliefs, and narrative goals. Interactive systems sustain SoA when their state transitions remain legible and responsive to user mental models~\citep{limerick2014experience, tchemeube2025evaluating}.

\paragraph{AI's Impact on Creative Agency, Cognitive Surrender, and Intent Debt.}

Recent empirical studies demonstrate that increasing AI automation can inadvertently degrade human SoA and psychological ownership across creative workflows, with novices and practicing creatives often experiencing steering and ownership differently~\citep{louie2020novice, krol2025exploring, krol2026supporting, brade2026agents}. \citet{he2026subjective} showed that automated assistance in creative tasks lowers creators' perceived ownership and belonging, an effect particularly pronounced among experienced practitioners. \citet{rafner2025agency} observed that SoA fluctuates throughout AI-assisted generation depending on how effectively users can steer AI outputs to match their original intent. At a foundational level, \citet{shaw2026thinking} proposed Tri-System Theory, extending dual-process models by introducing \emph{System~3} (Artificial Reasoning). Across preregistered experiments, they documented \emph{cognitive surrender}---the tendency for users to uncritically accept polished AI outputs, bypassing both intuitive evaluation (System~1) and deliberative reflection (System~2). Complementing this perspective, \citet{storey2026technical} highlighted how automated AI generation accumulates \emph{intent debt} and \emph{cognitive debt}, where the creator's underlying vision is progressively diluted by unexamined machine defaults, creating a disconnect between human intent and synthetic output.

A principal driver of cognitive surrender and intent debt is \emph{anchoring}~\citep{furnham2011literature}: once a user is presented with a high-fidelity rendered artifact, their cognitive search space contracts around the AI's generated choices, constraining subsequent iteration to reactive curation rather than open-ended correspondence~\citep{fernandes2025ai, ingold2013making}. To mitigate anchoring and prevent intent debt, \citet{shaw2026thinking} and \citet{storey2026technical} advocate for design interventions that surface assumptions, introduce productive conceptual reflection, and expose uncertainty indicators \emph{prior} to final artifact generation.

\paragraph{Shared Agency and Pre-Generation Control.}

As AI systems become more proactive, interaction design must evolve from opaque one-shot generation toward models of \emph{shared agency}~\citep{moruzzi2022creative, legaspi2024sense, rezwana2023designing}, where human and artificial agency are dynamically negotiated. However, most contemporary generative AI tools operate primarily on a \emph{post-generation critique} loop, positioning users to evaluate and edit finished outputs only after the model has silently resolved creative ambiguities. We argue that supporting shared agency in conceptual ideation and multi-modal prototyping requires shifting intervention upstream through \emph{pre-generation belief control}. By externalizing the AI's internal interpretation into an editable conceptual representation before rendering, systems can restore dialogic correspondence at the ideation stage and empower creators to establish the initial anchor.

\section{Detailed definitions and evaluation criteria for creative aspects}
\label{sec:appendix_creative_aspects}

To evaluate where in the workflow pre-generation belief control augments or diminishes human agency, we assess four core aspects of creative work, extending the classic analysis of creativity by~\citet{rhodes1961analysis}\footnote{These four aspects serve as a practical evaluation scaffold rather than an exhaustive taxonomy.}:
\begin{itemize}[leftmargin=*]
    \item \textbf{Intent:} Creators often approach initial subject matter to brainstorm, explore different ideas, and form their conceptual intent. We evaluate the extent to which an AI system augments or preserves the creator's original idea or concept, versus deflecting them from it or redirecting them toward repetitive, default realizations~\citep{doshi2024generative}.
    \item \textbf{Expression:} Creators use creative expression to explore stylistic trajectories suitable for their concept, applying personal artistic techniques, voice, and tone. We evaluate the extent to which the system encourages creators to articulate their stylistic choices and apply their own creative techniques, versus steering them toward homogeneous defaults~\citep{anderson2024homogenization, doshi2024generative}.
    \item \textbf{Process:} The non-linear progression in which creators move fluidly between exploration (where ideas and stylistic trajectories are incubated and rejected) and production (where ideas are finalized and concrete artifacts are refined)~\citep{botella2018stages}. We evaluate the extent to which the system allows creators to control how and when they explore alternatives, set priorities, and iterate on intermediate outcomes, versus curtailing their workflow~\citep{rezwana2023designing, yannakakis2014mixed}.
    \item \textbf{Outcome:} Concrete media artifacts produced during and at the conclusion of the creative process provide artistic direction and feedback until creators are satisfied with the final outcome. We evaluate the extent to which the system promotes a genuine sense of authorship over the final artifacts produced~\citep{epstein2020gets}.
\end{itemize}

\section{Implementation details}
\label{ssec:compute}

PaCC is implemented as a single-page web application built with React~19, TypeScript, and Vite, using the \texttt{@google/genai} JavaScript SDK to communicate with Google's Gemini APIs. The front-end comprises a main application controller, interactive components for the belief graph, clarification cards, attribute inspector, and prompt review modal, and a centralized service module that orchestrates all LLM calls. The source code is publicly available.\footnote{\url{https://aistudio.google.com/apps/bundled/proactive_co_creator?showCode=true}}

\paragraph{LLM call pipeline.} Each analysis cycle in PaCC issues three LLM calls, each with a distinct system prompt and a typed JSON response schema enforced via the Gemini API's \texttt{responseSchema} parameter (ensuring outputs conform to the expected entity--attribute--relationship structure without post-hoc parsing):

\begin{enumerate}
    \item \textbf{Belief graph and media attributes} (\texttt{parsePromptToBeliefGraph}). Given the user's prompt, optional reference images, the selected modality (image/story/video), and the proactiveness tier, this call produces a structured JSON object containing: (a)~\emph{entities} with names, descriptions, explicit/implicit provenance flags, and per-entity attributes each with 2--3 candidate values; and (b)~\emph{relationships} between entities (spatial, emotional, functional), also marked as explicit or inferred. Mode-specific instructions inject mandatory scene-level attributes (e.g., camera angle and lighting for images; pacing and duration for videos).
    \item \textbf{Clarification generation} (\texttt{generateClarifications}). Issued in parallel with the belief graph call, this produces exactly three atomic clarifying questions, each with 3--5 selectable options. The system prompt conditions on the proactiveness tier, the current prompt, any reference images, and previously asked questions to avoid repetition.
    \item \textbf{Prompt refinement} (\texttt{refinePromptWithAllUpdates}). After the user answers clarification questions and/or edits the belief graph, this call integrates all updates, including question-answer pairs and graph edits (attribute changes, entity additions/removals, relationship modifications), into a single refined prompt. The model is instructed to preserve all existing details not explicitly changed, remove entities marked absent, and respect ``You Decide'' (AI delegation) selections (temperature~$= 0.2$).
\end{enumerate}

\paragraph{Incremental graph updates.} When the user modifies the prompt after the initial belief graph has been constructed, PaCC issues an incremental \texttt{updateBeliefGraph} call that produces a \emph{JSON delta patch} rather than regenerating the full graph. The delta schema supports six operations: add/remove/update entities (including entity renaming and attribute add/remove) and add/remove/update relationships. The client applies these patches locally, preserving entity identity and relationship continuity across edits. If the incremental update fails, the system falls back to full graph regeneration.

\paragraph{Proactiveness tiers.} PaCC supports three proactiveness levels (\emph{low}, \emph{medium}, and \emph{high}) that modulate both the belief graph and clarification system prompts. At \emph{low}, the model extracts only entities and attributes explicitly mentioned in the prompt. At \emph{medium}, it infers reasonable defaults for missing attributes and may add 1--2 implicit entities. At \emph{high}, it proactively invents creative attributes, generates 3+ implicit entities, and suggests novel stylistic or narrative elements. Users can also delegate individual clarification questions to the AI via the ``You Decide'' button.

\paragraph{Reliability.} All API calls use exponential-backoff retry (up to 5 attempts with jittered delays) for transient errors (HTTP 429, 500, 502, 503, 504). If the user-selected model fails after retries, the system falls back to \texttt{gemini-3-flash-preview}.

\paragraph{Models.}
PaCC was built using the app building functionality in Google AI Studio (\url{https://aistudio.google.com/apps}), with the application code itself authored with the assistance of Gemini models within the same environment. PaCC is configured to use Gemini APIs (\url{https://ai.google.dev/gemini-api/docs/models}). Text models (belief graph, clarifications, prompt refinement, story generation): gemini-3.1-pro-preview, gemini-3-flash-preview, gemini-3.1-flash-lite-preview, gemini-flash-latest. Image generation: gemini-2.5-flash-image, gemini-3.1-flash-image-preview, gemini-3-pro-image-preview, imagen-4.0-generate-001. Video generation: veo-3.1-fast-generate-preview, veo-3.1-generate-preview. Participants used the free tier for text and image generation and tier~1 for video generation.

\begin{table}[h]
\centering
\caption{Mapping of PaCC features to design principles. DP1: Intervene before generation. DP2: Make assumptions visible. DP3: Preserve the creative possibility space. DP4: Calibrate AI autonomy.}
\label{tab:principle_feature_map}
\small
\begin{tabular}{@{}llcccc@{}}
\toprule
Feature & Mechanism & DP1 & DP2 & DP3 & DP4 \\
\midrule
\multicolumn{6}{@{}l}{\emph{Pre-generation Control Features}} \\
Clarifying Questions   & Targeted Q\&A with selectable options & \checkmark & \checkmark & \checkmark & \checkmark \\
Belief Graph           & Editable entity--relationship graph   & \checkmark & \checkmark & \checkmark & \checkmark \\
Attributes             & Scene-level parameters with candidates & \checkmark & \checkmark & \checkmark & \checkmark \\
\addlinespace
\multicolumn{6}{@{}l}{\emph{Prompt Refinement Capabilities}} \\
Refined Prompt Review  & Original vs.\ editable refined prompt & \checkmark & \checkmark &  &  \\
Proactiveness Tiers    & Low/medium/high AI initiative control &  &  &  & \checkmark \\
``You Decide''         & Per-question delegation to the AI     &  &  &  & \checkmark \\
\bottomrule
\end{tabular}
\end{table}

\subsection{Example User Journey and Pro Workflow}
\label{ssec:user_journey}

To demonstrate how PaCC enables a progressive, multi-modal ``pro workflow,'' Figures~\ref{fig:user_journey_stage1_a}--\ref{fig:user_journey_stage3} walk through an end-to-end example user journey spanning Image Mode, Story Mode, and Video Generation Mode.

\subsubsection{Stage 1: Image Mode (Concept Visualization)}
\begin{figure}[htbp]
    \centering
    \begin{subfigure}[b]{0.9\textwidth}
        \includegraphics[width=\textwidth]{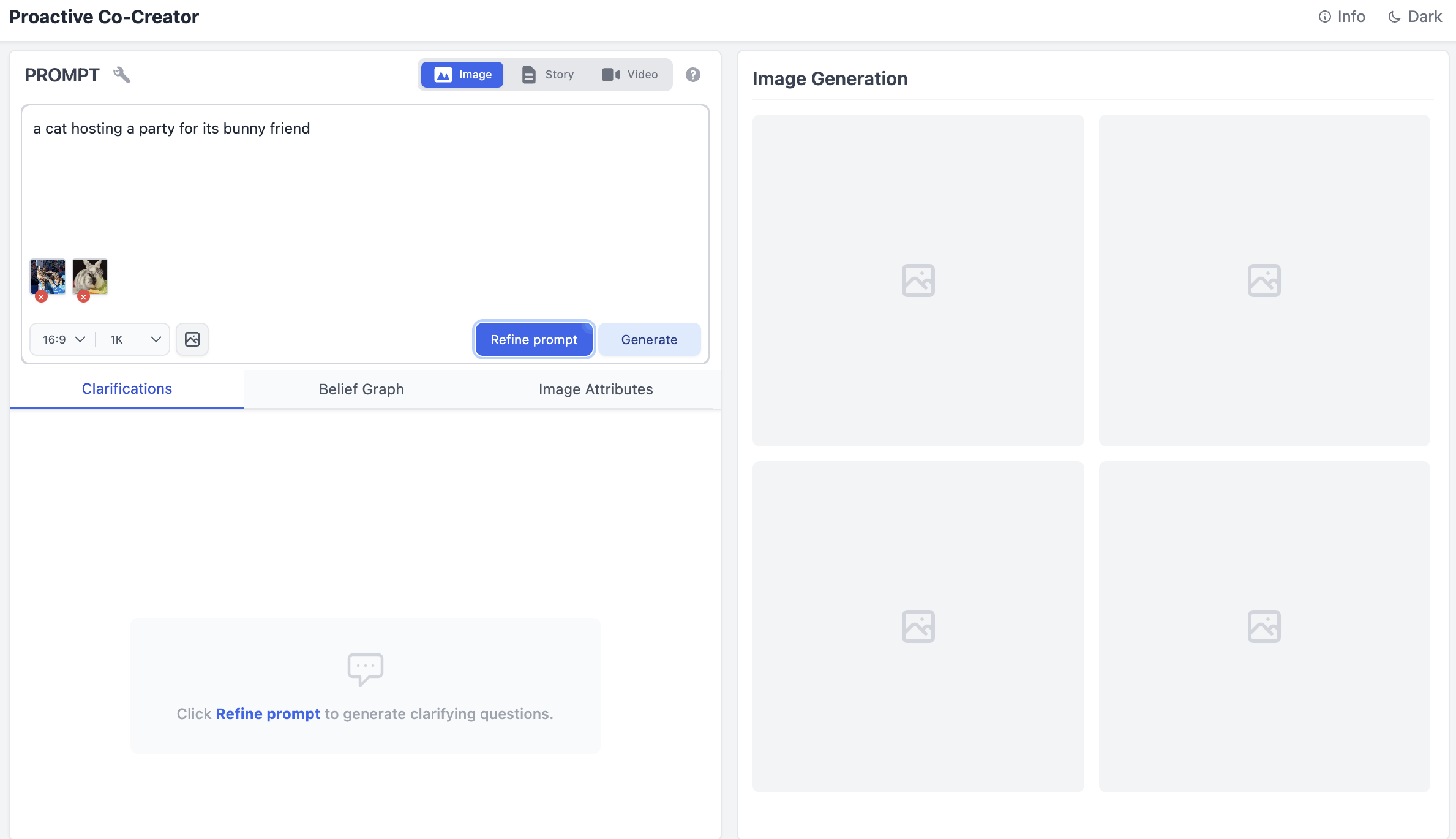}
        \caption{Step 1: Initial prompt entry in Image Mode.}
        \label{fig:step1}
    \end{subfigure}
    
    \vspace{0.5em}
    
    \begin{subfigure}[b]{0.9\textwidth}
        \includegraphics[width=\textwidth]{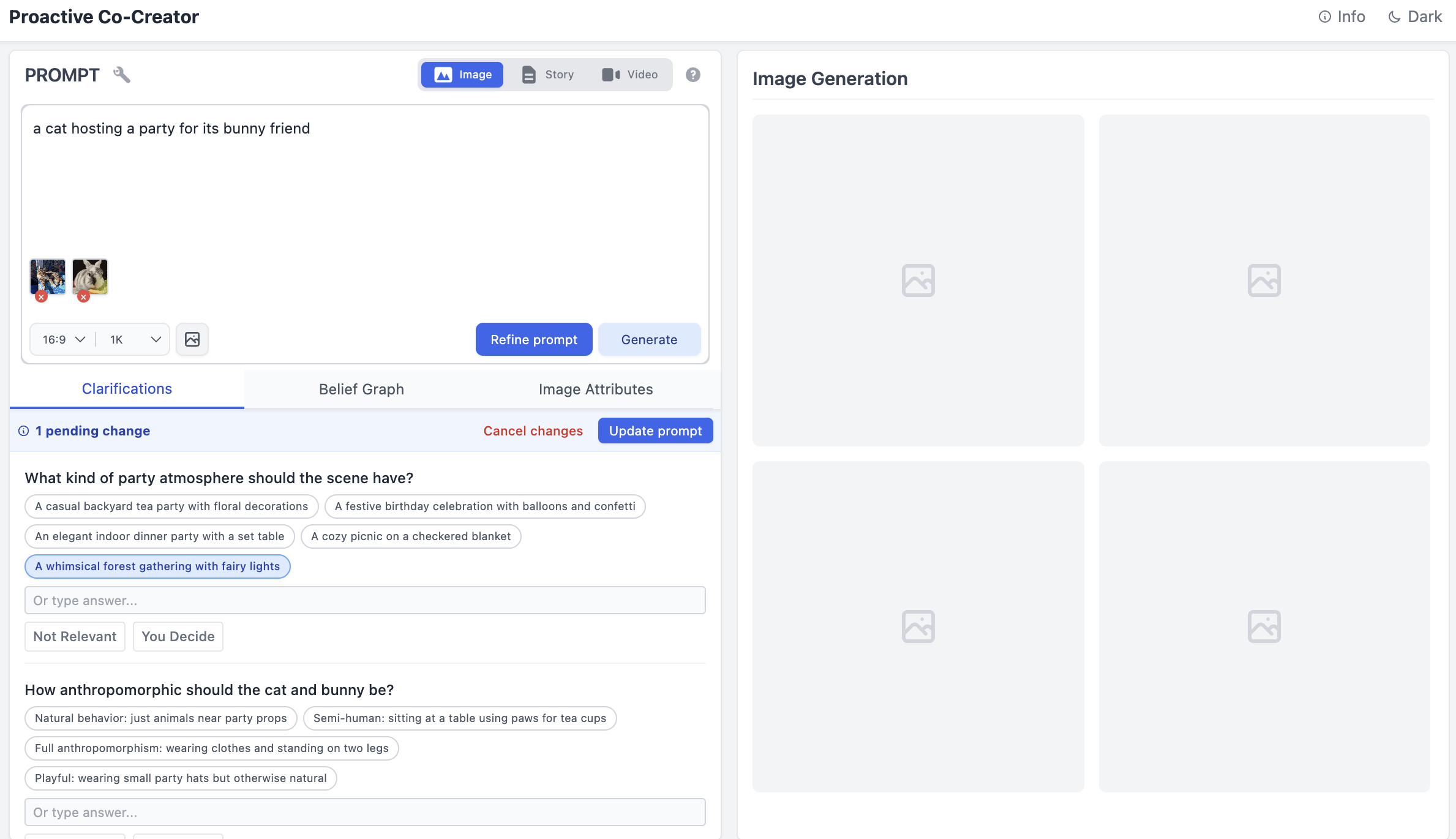}
        \caption{Step 2: Answering pre-generation clarifying questions.}
        \label{fig:step2}
    \end{subfigure}
    \caption{Stage 1 (Image Mode): Visualizing the concept and exercising pre-generation belief control (Steps 1--2).}
    \label{fig:user_journey_stage1_a}
\end{figure}

\begin{figure}[htbp]
    \centering
    \begin{subfigure}[b]{0.9\textwidth}
        \includegraphics[width=\textwidth]{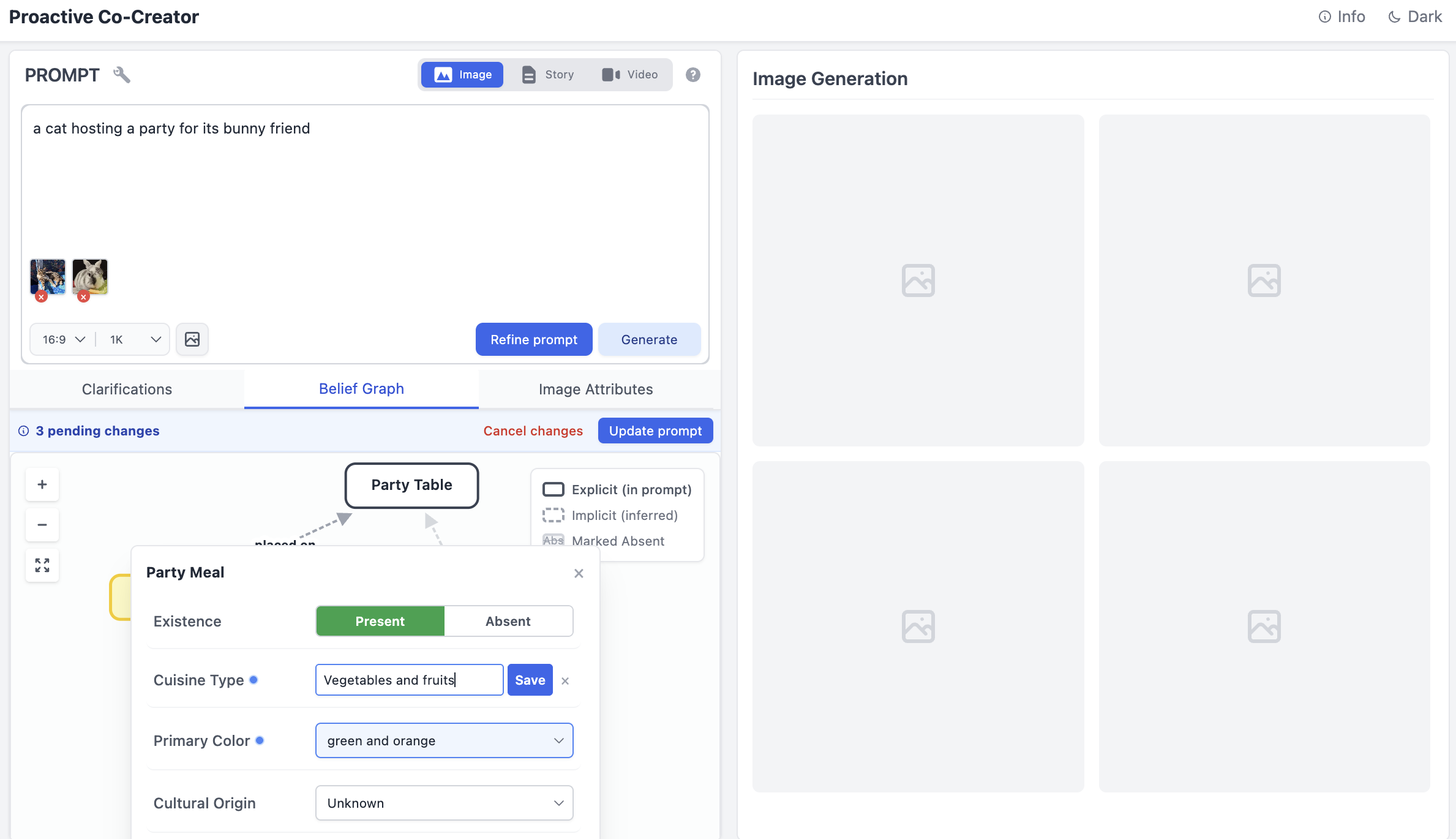}
        \caption{Step 3: Inspecting and editing the belief graph.}
        \label{fig:step3}
    \end{subfigure}
    
    \vspace{0.5em}
    
    \begin{subfigure}[b]{0.9\textwidth}
        \includegraphics[width=\textwidth]{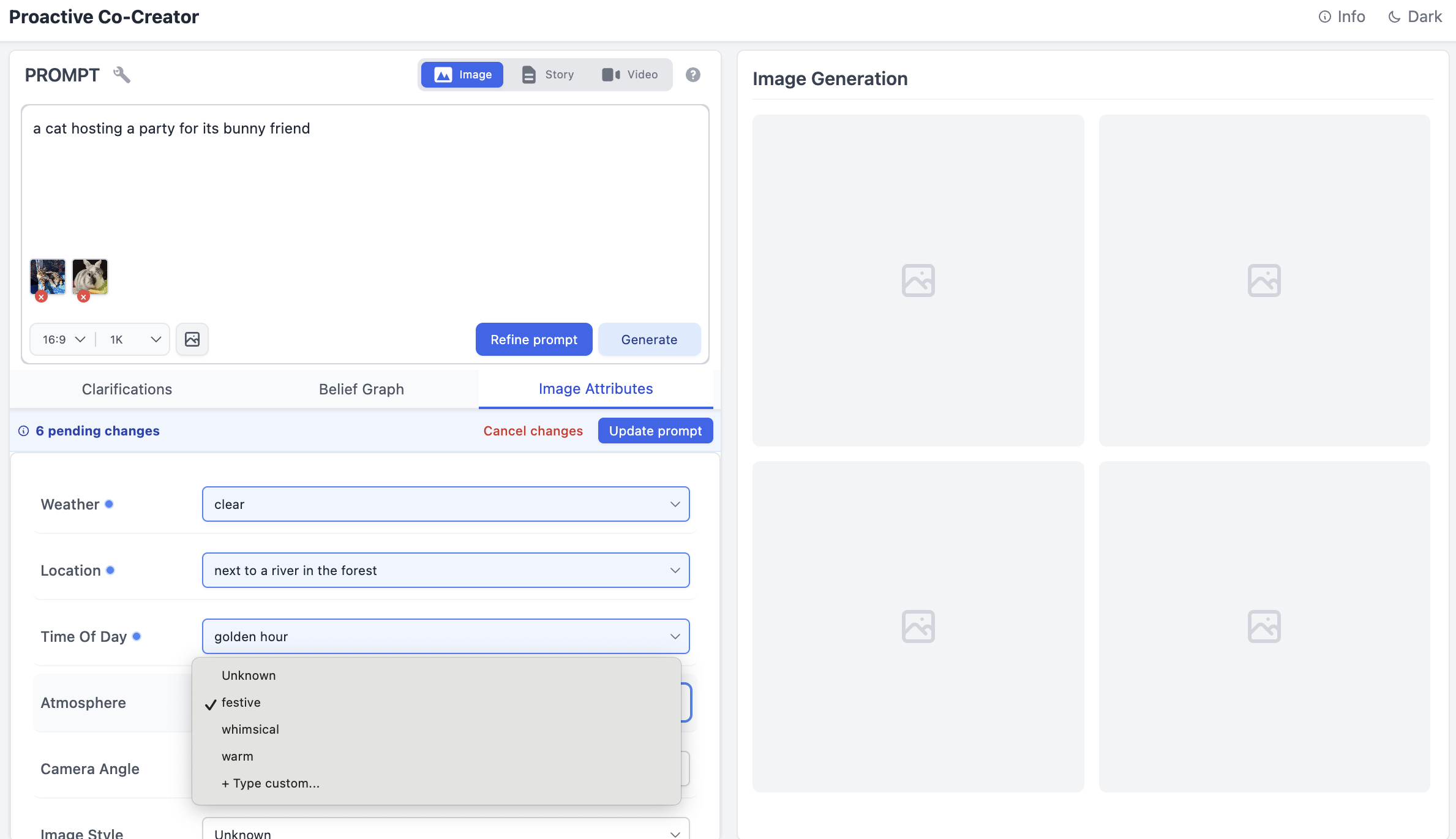}
        \caption{Step 4: Adjusting scene-level attributes.}
        \label{fig:step4}
    \end{subfigure}
    \caption{Stage 1 (Image Mode): Inspecting belief graph and adjusting attributes (Steps 3--4).}
    \label{fig:user_journey_stage1_b}
\end{figure}

\begin{figure}[htbp]
    \centering
    \begin{subfigure}[b]{0.9\textwidth}
        \includegraphics[width=\textwidth]{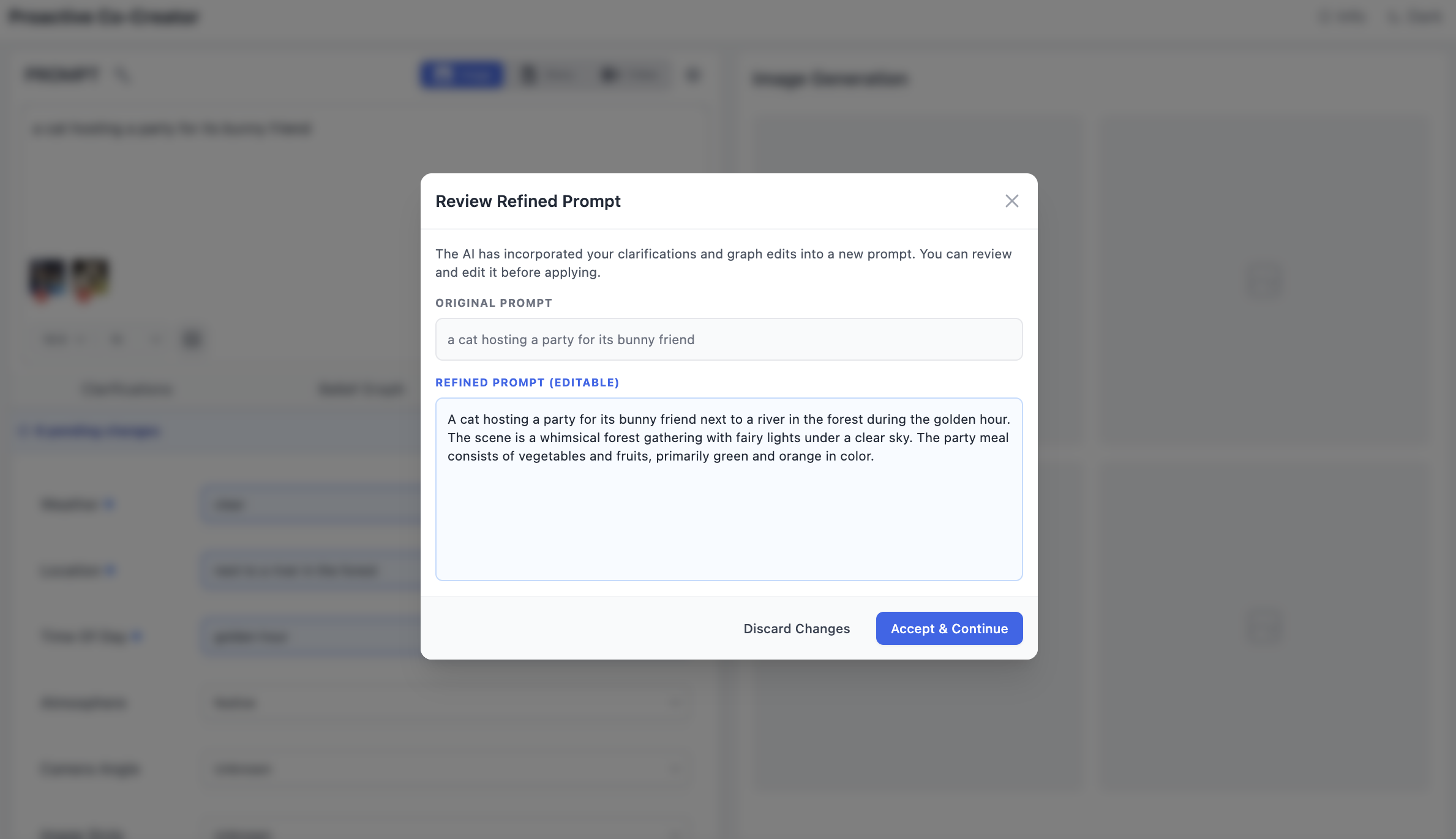}
        \caption{Step 5: Reviewing the side-by-side prompt diff.}
        \label{fig:step5}
    \end{subfigure}
    
    \vspace{0.5em}
    
    \begin{subfigure}[b]{0.9\textwidth}
        \includegraphics[width=\textwidth]{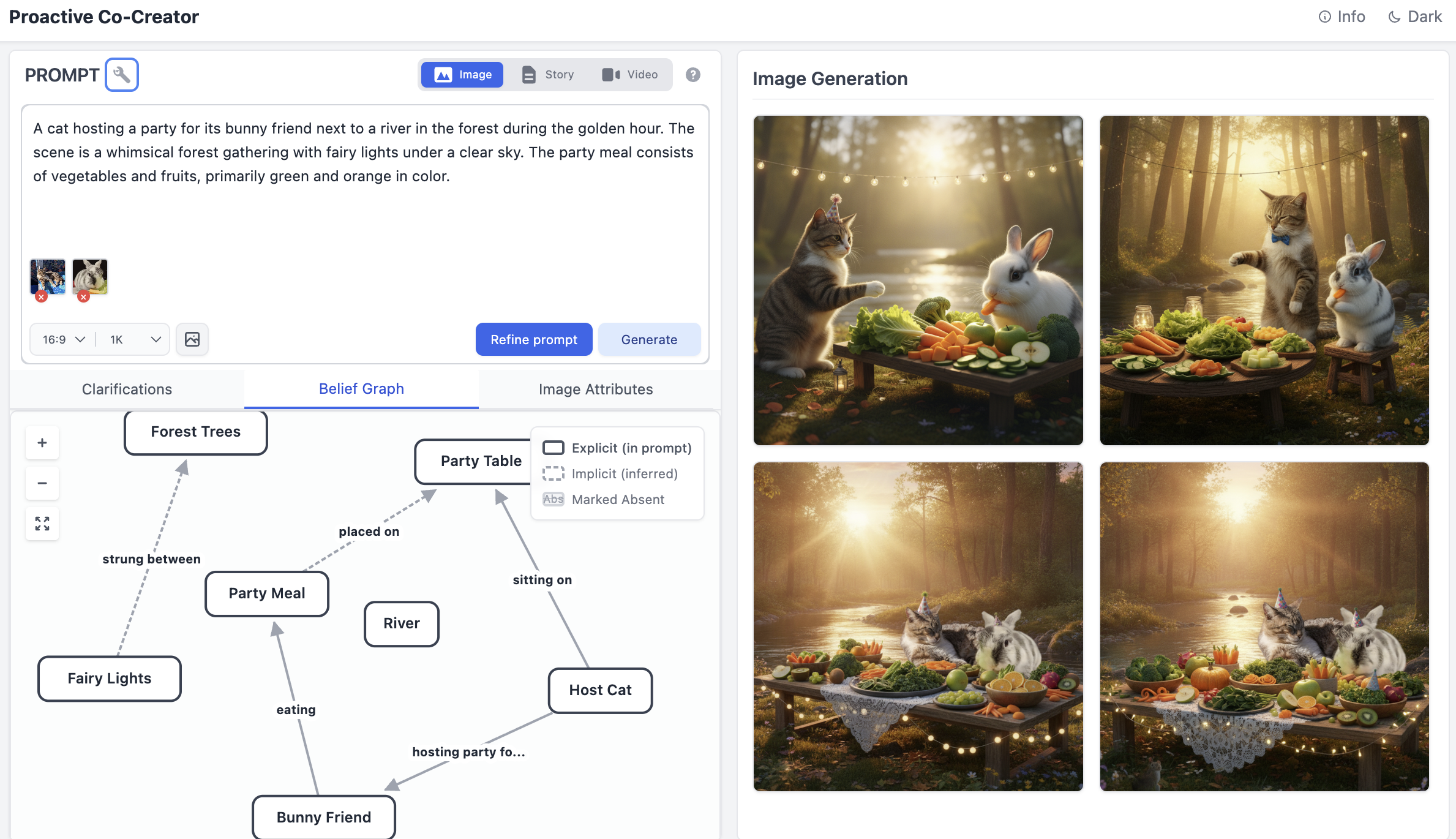}
        \caption{Step 6: Reviewing generated image candidates.}
        \label{fig:step6}
    \end{subfigure}
    \caption{Stage 1 (Image Mode): Reviewing prompt diff and generating images (Steps 5--6).}
    \label{fig:user_journey_stage1_c}
\end{figure}

\begin{figure}[htbp]
    \centering
    \begin{subfigure}[b]{0.9\textwidth}
        \includegraphics[width=\textwidth]{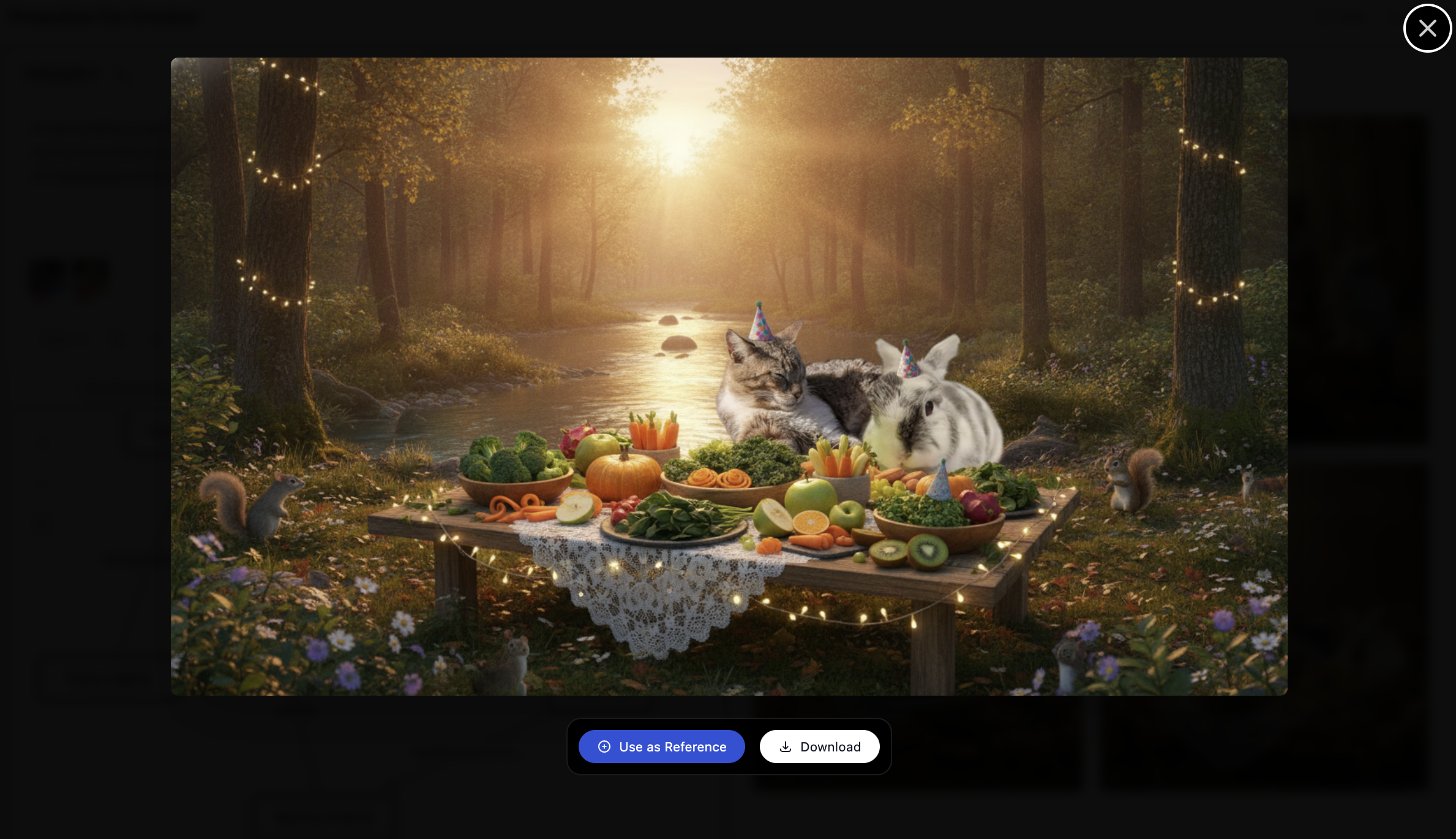}
        \caption{Step 7: Selecting an image and clicking ``Use as Reference''.}
        \label{fig:step7}
    \end{subfigure}
    \caption{Stage 1 (Image Mode): Selecting a visual reference image (Step 7).}
    \label{fig:user_journey_stage1_d}
\end{figure}

The creator begins by visualizing their concept, applying pre-generation belief controls, generating candidate images, and selecting a key result as a visual reference (Figures~\ref{fig:user_journey_stage1_a}--\ref{fig:user_journey_stage1_d}):
\begin{enumerate}[leftmargin=*]
    \item \textbf{Initial Prompt Entry:} The user types an initial creative prompt (e.g., ``a cat hosting a party for its bunny friends'') in Image Mode (Figure~\ref{fig:step1}).
    \item \textbf{Clarification Questions:} PaCC surfaces targeted pre-generation questions to clarify ambiguous scene attributes (Figure~\ref{fig:step2}).
    \item \textbf{Belief Graph Inspection:} The user inspects and updates the structured belief graph, distinguishing explicit prompt entities from AI-inferred ones (Figure~\ref{fig:step3}).
    \item \textbf{Attribute Calibration:} Output-level parameters like lighting, camera angle, and artistic style are adjusted in the attributes panel (Figure~\ref{fig:step4}).
    \item \textbf{Prompt Review:} The user compares the original and AI-enriched prompts side by side before generation (Figure~\ref{fig:step5}).
    \item \textbf{Image Generation:} PaCC generates high-fidelity candidate images reflecting the refined belief graph (Figure~\ref{fig:step6}).
    \item \textbf{Reference Image Selection:} The user selects their preferred image and clicks ``Use as Reference'' to lock the visual baseline for multi-modal expansion (Figure~\ref{fig:step7}).
\end{enumerate}

\subsubsection{Stage 2: Story Mode (Cinematic Video Description)}
\begin{figure}[htbp]
    \centering
    \begin{subfigure}[b]{0.9\textwidth}
        \includegraphics[width=\textwidth]{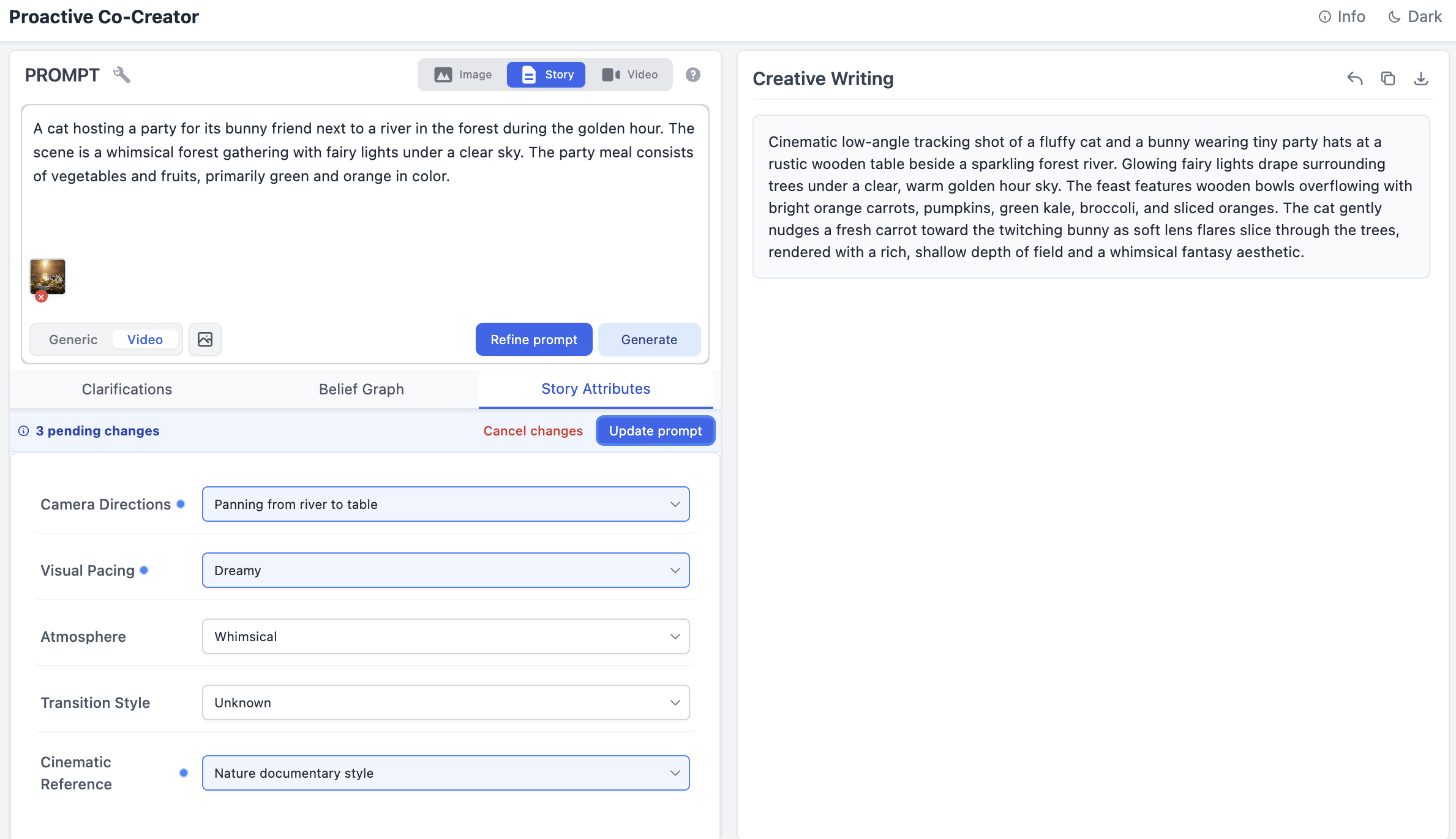}
        \caption{Step 8: Switching to Video settings and drafting scene description.}
        \label{fig:step8}
    \end{subfigure}
    
    \vspace{0.6em}
    
    \begin{subfigure}[b]{0.9\textwidth}
        \includegraphics[width=\textwidth]{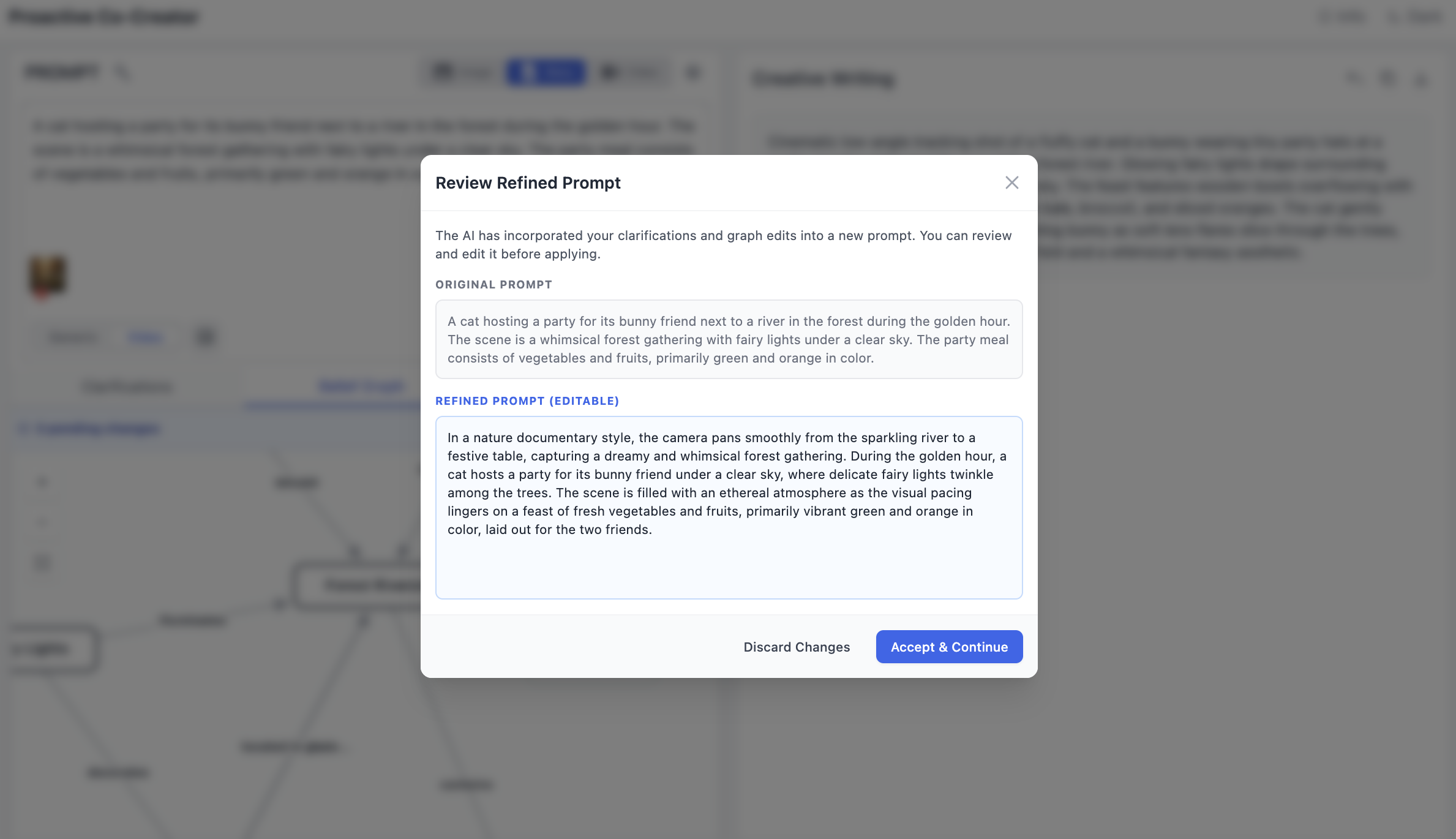}
        \caption{Step 9: Generating and copying the refined cinematic video prompt.}
        \label{fig:step9}
    \end{subfigure}
    \caption{Stage 2 (Story Mode / Video Description): Expanding the concept into an enriched cinematic video description.}
    \label{fig:user_journey_stage2}
\end{figure}

The creator expands the static concept into a temporal, cinematic narrative (Figure~\ref{fig:user_journey_stage2}):
\begin{enumerate}[leftmargin=*]
    \item[8.] \textbf{Video Description Setup:} The user switches to ``Video'' mode in settings and provides high-level motion and plot directions (Figure~\ref{fig:step8}).
    \item[9.] \textbf{Cinematic Prompt Generation:} PaCC generates an enriched cinematic video description detailing camera movement, atmospheric pacing, and character actions, which the user copies for video generation (Figure~\ref{fig:step9}).
\end{enumerate}

\subsubsection{Stage 3: Video Mode (Multi-Modal Generation)}
\begin{figure}[htbp]
    \centering
    \includegraphics[width=0.9\textwidth]{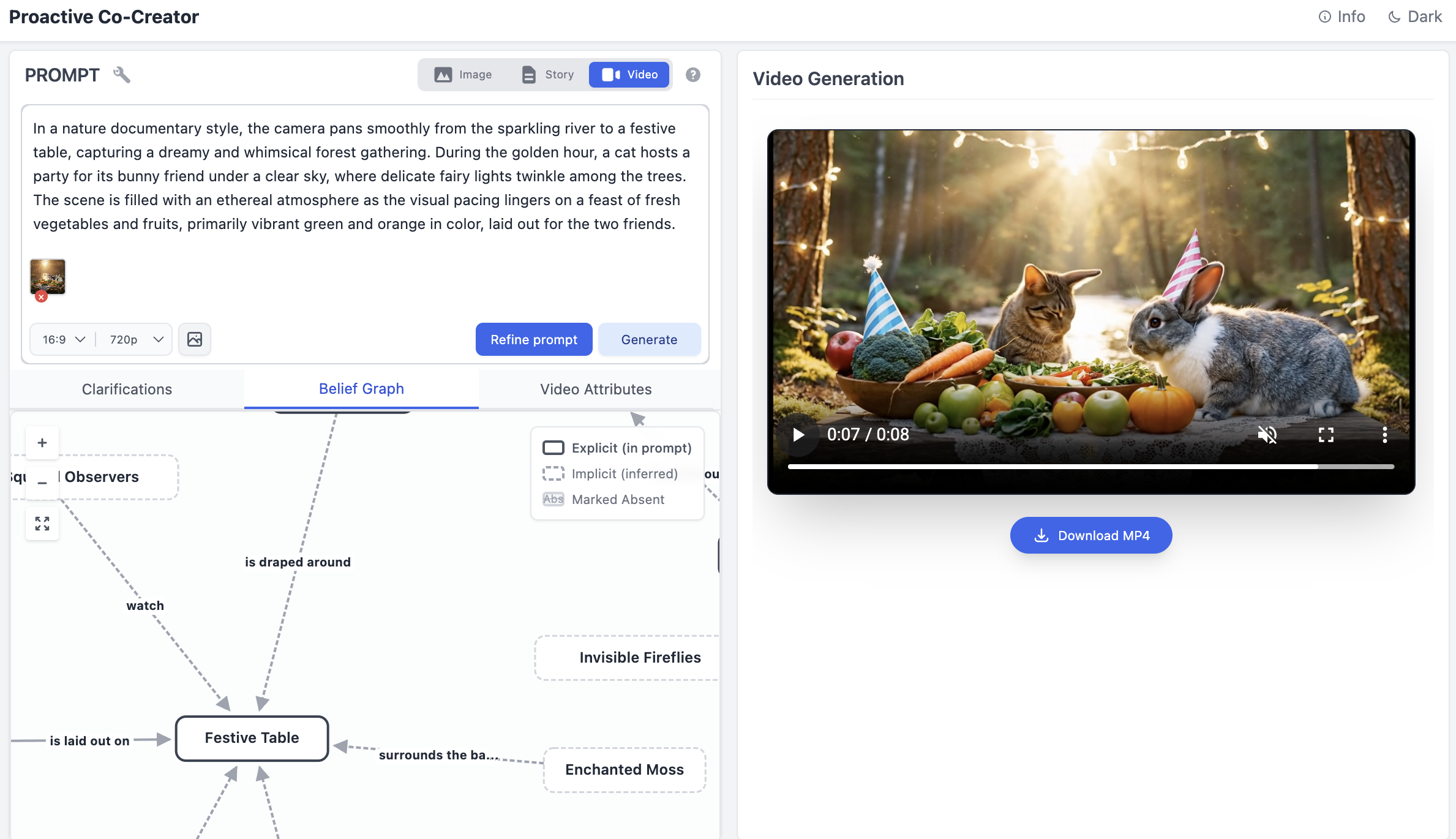}
    \caption{Stage 3 (Video Mode): Step 10: Combining the reference image and cinematic video description to generate the final video.}
    \label{fig:user_journey_stage3}
    \label{fig:step10}
\end{figure}

The creator can continue the prompt refinement using the pre-generation control in the video mode. Here, the creator generates the final dynamic video artifact (Figure~\ref{fig:user_journey_stage3}):
\begin{enumerate}[leftmargin=*]
    \item[10.] \textbf{Video Generation:} The user pastes the refined cinematic description, links the reference image selected in Stage 1, and generates the final video output (Figure~\ref{fig:step10}).
\end{enumerate}

\section{Details on human studies}
\label{sec:appendix_human_studies}

\subsection{Extended study design, participant demographics, and procedure}
\label{sec:appendix_study_details}

\paragraph{Participants.} We recruited 8 participants (2 female, 6 male) from our institution, working in AI or AI-related professions spanning research, design, and engineering roles from locations including New York, Boston, Maryland, San Francisco, Mountain View, and Seattle (see Table~\ref{tab:participants}). We excluded AI artists (individuals who use AI as their primary medium for commercial art creation), as that population would require a different study setup. Participants received a \$40 gift card for completing the study. 5 of 8 participants worked in foundational or applied AI development and research, 2 in creative roles, and 1 in a non-creative professional role. All participants reported using AI in the past 6 months, during which 7 participants used AI to complete work projects, 5 to explore ideas for work, 5 to complete personal art projects, and 5 to explore ideas for personal projects.

\paragraph{Procedure.} Following a 10-minute demonstration of PaCC's features (Clarifying Questions, Belief Graph, Attribute Choices) for interactive prompt refinement, participants spent 30 minutes on any unconstrained media generation task that they considered artistic or creative with no completion requirements. Participants were encouraged to use all pre-generation controls across writing, image generation, and video generation with no further instructions. Authors assisted with technical issues, of which three arose: participants required API keys, model rate limits prompted participants to switch models or modality during the study, and safety filters occasionally required prompt adjustments. Participants were then introduced to the study goals (\textit{"evaluating how PaCC affects an individual's SoA"}) and completed a 20-minute Google Form survey (Appendix~\ref{sec:user_study_form}). The survey captured participants' job titles, job descriptions, and use of generative AI in the last 6 months. The next section asked for a definition of SoA and ratings of whether the participants' current AI tools augmented, preserved, or diminished their SoA. The third section captured whether PaCC decreased or increased their SoA on a 5-point Likert scale (1=Diminished, 3=Preserved, 5=Augmented). Further sections for individual PaCC features used similar five-point items for rating sense of agency, ownership, control and authorship across creative intent, expression, process, and outcomes. The last section asked participants to rate whether the sense of agency, ownership, control, and authorship experienced with PaCC was greater than, equal to, or less than that of their current tools. Each survey section closed with an optional written explanation and all survey questions were optional. The total study time was 60 minutes, and participants who completed the survey early engaged in a semi-structured discussion. Participant risks were minimal and consistent with using an interface and completing a survey. Additional risks from exposure to AI systems were mitigated by safety policies which prevented users from engaging with harmful outputs.

\definecolor{rowlight}{gray}{0.95}

\begin{table*}[htbp]
\centering
\caption{Participant demographics, professional roles, prior AI co-creation experience, and self-reported definitions of sense of agency. Participants span AI research (5), creative design (2), and non-creative professional (1) backgrounds.}
\label{tab:participants}
\small
\renewcommand{\arraystretch}{1.35}
\begin{tabularx}{\textwidth}{@{} c c l l c X @{}}
\toprule
\textbf{ID} & \textbf{Gender} & \textbf{Occupation} & \textbf{Professional Role} & \textbf{AI Exp.} & \textbf{Self-Reported Definition of Agency} \\
\midrule
\rowcolor{rowlight}
P1 & M & Research Engineer & AI Professional & \checkmark\checkmark\checkmark\checkmark &
The ability to influence the final output of a project or effort I'm working on. \\
P2 & M & Research Scientist & AI Professional & \checkmark\checkmark\checkmark\checkmark &
How effectively I can communicate my intent to the AI and have it reliably follow my instructions. \\
\rowcolor{rowlight}
P3 & M & Student Researcher & AI Professional & \checkmark &
I feel that I am the initiator of my actions. \\
P4 & M & UX Researcher & Other Professional & \checkmark\checkmark\checkmark\checkmark &
Control in the creation process. \\
\rowcolor{rowlight}
P5 & F & Content Designer & Creative Professional & \checkmark\checkmark &
Feeling like I can take some credit for the output, that the output was based enough on my decisions to feel like mine. \\
P6 & F & UX Researcher & AI Professional & \checkmark &
Having informed choices to make. \\
\rowcolor{rowlight}
P7 & M & Sr.\ Interaction Designer & Creative Professional & \checkmark\checkmark &
Freedom to express my thoughts without friction and feeling heard. \\
P8 & M & Software Engineer & AI Professional & \checkmark\checkmark\checkmark\checkmark &
Feels like I'm in charge of what I'm building out, especially with AI tools. \\
\bottomrule
\end{tabularx}
\vspace{2pt}
\parbox{\textwidth}{\footnotesize
\textbf{AI Exp.}\ indicates the breadth of prior media-generation AI experience in the past 6 months. Each \checkmark\ represents one type of activity from: (1)~completed a personal art project, (2)~completed a work project, (3)~explored ideas for a personal project, (4)~explored ideas for a work project. \textbf{Professional Role} categories: \emph{AI Professional} = foundational or applied AI R\&D; \emph{Creative Professional} = creative role (e.g., UX, VisD, Motion Design); \emph{Other Professional} = non-creative role using AI to supplement work.}
\end{table*}

In Table~\ref{tab:participants}, we include the anonymized details of the participants.

\subsection{Per-feature Likert ratings}
\label{sec:appendix_per_feature}

Tables~\ref{tab:cq_detail}--\ref{tab:pr_detail} report the mean Likert ratings (1--5 scale) for each PaCC feature, broken down by creative aspect and agency dimension. Scores above~3 indicate that the feature increased or augmented the dimension relative to participants' baseline experience.

\begin{table}[h]
\centering
\caption{Clarifying Questions: Mean Likert ratings by creative aspect and agency dimension.}
\label{tab:cq_detail}
\small
\begin{tabular}{lcccc}
\toprule
Aspect & Agency & Ownership & Control & Authorship \\
\midrule
Intent      & $3.88 \pm 0.64$ & $3.25 \pm 0.89$ & $3.62 \pm 0.92$ & $3.29 \pm 0.76$ \\
Expression  & $3.50 \pm 0.53$ & $3.25 \pm 0.71$ & $3.50 \pm 0.76$ & $3.14 \pm 0.69$ \\
Process     & $3.62 \pm 0.92$ & $3.38 \pm 0.92$ & $3.75 \pm 0.89$ & $3.43 \pm 0.98$ \\
Outcome     & $3.62 \pm 0.92$ & $3.50 \pm 0.93$ & $3.75 \pm 0.89$ & $2.86 \pm 0.90$ \\
\bottomrule
\end{tabular}
\end{table}

\begin{table}[h]
\centering
\caption{Belief Graph: Mean Likert ratings by creative aspect and agency dimension ($N{=}7$; one participant did not use this feature).}
\label{tab:bg_detail}
\small
\begin{tabular}{lcccc}
\toprule
Aspect & Agency & Ownership & Control & Authorship \\
\midrule
Intent      & $3.71 \pm 0.95$ & $3.43 \pm 0.79$ & $3.43 \pm 0.98$ & $3.33 \pm 0.52$ \\
Expression  & $3.71 \pm 1.11$ & $3.43 \pm 0.98$ & $3.57 \pm 0.98$ & $3.00 \pm 0.89$ \\
Process     & $3.86 \pm 0.90$ & $3.57 \pm 0.79$ & $3.71 \pm 1.11$ & $3.00 \pm 0.89$ \\
Outcome     & $3.71 \pm 0.95$ & $3.43 \pm 0.79$ & $3.43 \pm 0.98$ & $3.00 \pm 0.89$ \\
\bottomrule
\end{tabular}
\end{table}

\begin{table}[h]
\centering
\caption{Attributes: Mean Likert ratings by creative aspect and agency dimension ($N{=}7$; one participant did not use this feature).}
\label{tab:attr_detail}
\small
\begin{tabular}{lcccc}
\toprule
Aspect & Agency & Ownership & Control & Authorship \\
\midrule
Intent      & $3.29 \pm 0.95$ & $3.29 \pm 0.95$ & $3.57 \pm 0.79$ & $3.17 \pm 0.75$ \\
Expression  & $3.29 \pm 0.76$ & $3.14 \pm 0.69$ & $3.43 \pm 0.53$ & $3.00 \pm 0.63$ \\
Process     & $3.29 \pm 0.76$ & $3.29 \pm 0.76$ & $3.29 \pm 0.49$ & $3.00 \pm 0.63$ \\
Outcome     & $3.29 \pm 0.95$ & $3.43 \pm 0.98$ & $3.43 \pm 0.79$ & $3.00 \pm 0.63$ \\
\bottomrule
\end{tabular}
\end{table}

\begin{table}[h]
\centering
\caption{Prompt Refinement: Mean Likert ratings by creative aspect and agency dimension.}
\label{tab:pr_detail}
\small
\begin{tabular}{lcccc}
\toprule
Aspect & Agency & Ownership & Control & Authorship \\
\midrule
Intent      & $4.00 \pm 1.07$ & $3.62 \pm 1.19$ & $3.88 \pm 1.36$ & $3.43 \pm 1.40$ \\
Expression  & $3.75 \pm 1.04$ & $3.50 \pm 1.20$ & $3.62 \pm 1.30$ & $3.29 \pm 1.38$ \\
Process     & $3.75 \pm 1.04$ & $3.50 \pm 1.20$ & $3.62 \pm 1.30$ & $3.29 \pm 1.38$ \\
Outcome     & $3.88 \pm 1.13$ & $3.50 \pm 1.41$ & $3.75 \pm 1.39$ & $3.29 \pm 1.38$ \\
\bottomrule
\end{tabular}
\end{table}

\subsection{Baseline AI tool assessment}
\label{sec:appendix_baseline}

Before using PaCC, participants reported on whether their current AI tools preserve, augment, or diminish their sense of agency. The distribution of responses is shown in Table~\ref{tab:baseline_ai}. The most common response was ``Sometimes'' for both preserving and diminishing agency, indicating that participants had mixed prior experiences with AI tools and did not uniformly perceive current tools as either beneficial or harmful to their agency.

\begin{table}[h]
\centering
\caption{Participants' baseline assessment of current AI tools' impact on their sense of agency ($N{=}8$).}
\label{tab:baseline_ai}
\small
\begin{tabular}{lccccc}
\toprule
Statement & Always & Very often & Sometimes & Rarely & Never \\
\midrule
AI tools preserve my agency  & 0 & 2 & 5 & 1 & 0 \\
AI tools augment my agency   & 0 & 2 & 4 & 2 & 0 \\
AI tools diminish my agency  & 0 & 1 & 7 & 0 & 0 \\
\bottomrule
\end{tabular}
\end{table}

\subsection{Feature polarization analysis}
\label{sec:appendix_polarization}

Table~\ref{tab:polarization} reports the standard deviation and range of per-participant mean ratings for each feature across agency dimensions, indicating the degree of consensus or disagreement. Ownership was the most polarizing dimension overall ($\text{SD}$ up to $1.49$ for PaCC overall), while clarifying questions and media attributes showed the strongest consensus (average $\text{SD} < 0.75$). Prompt refinement had the highest overall mean but also the most disagreement.

\begin{table}[h]
\centering
\caption{Polarization of participant ratings: SD and range of per-participant mean ratings by feature and dimension. Higher SD indicates more disagreement among participants.}
\label{tab:polarization}
\small
\begin{tabular}{llcccc}
\toprule
Feature & Dimension & Mean & SD & Range & $n$ \\
\midrule
\rowcolor{rowlight}
PaCC (overall)       & Agency     & 4.09 & 0.84 & 2.5--5.0 & 8 \\
PaCC (overall)       & Ownership  & 3.22 & \textbf{1.49} & 1.2--5.0 & 8 \\
\rowcolor{rowlight}
PaCC (overall)       & Control    & 3.89 & 0.69 & 3.0--5.0 & 7 \\
PaCC (overall)       & Authorship & 2.75 & 1.13 & 1.5--5.0 & 7 \\
\midrule
\rowcolor{rowlight}
Clarifying Questions & Agency     & 3.66 & 0.64 & 2.8--4.5 & 8 \\
Clarifying Questions & Ownership  & 3.34 & 0.77 & 2.0--4.5 & 8 \\
\rowcolor{rowlight}
Clarifying Questions & Control    & 3.66 & 0.79 & 2.0--4.5 & 8 \\
Clarifying Questions & Authorship & 3.18 & 0.72 & 2.0--4.0 & 7 \\
\midrule
\rowcolor{rowlight}
Belief Graph         & Agency     & 3.75 & 0.95 & 2.8--5.0 & 7 \\
Belief Graph         & Ownership  & 3.46 & 0.80 & 2.8--5.0 & 7 \\
\rowcolor{rowlight}
Belief Graph         & Control    & 3.54 & 0.98 & 2.0--5.0 & 7 \\
Belief Graph         & Authorship & 3.08 & 0.79 & 2.2--4.0 & 6 \\
\midrule
\rowcolor{rowlight}
Attributes           & Agency     & 3.29 & 0.81 & 2.0--4.5 & 7 \\
Attributes           & Ownership  & 3.29 & 0.81 & 2.0--4.5 & 7 \\
\rowcolor{rowlight}
Attributes           & Control    & 3.43 & 0.61 & 3.0--4.5 & 7 \\
Attributes           & Authorship & 3.04 & 0.64 & 2.0--4.0 & 6 \\
\midrule
\rowcolor{rowlight}
Prompt Refinement    & Agency     & 3.84 & 1.04 & 2.0--5.0 & 8 \\
Prompt Refinement    & Ownership  & 3.53 & 1.21 & 2.0--5.0 & 8 \\
\rowcolor{rowlight}
Prompt Refinement    & Control    & 3.72 & 1.32 & 1.0--5.0 & 8 \\
Prompt Refinement    & Authorship & 3.32 & \textbf{1.37} & 1.0--5.0 & 7 \\
\bottomrule
\end{tabular}
\end{table}

\subsection{Qualitative responses: Most and least useful features}
\label{sec:appendix_qualitative}

Table~\ref{tab:most_least_useful} summarizes participants' responses when asked which PaCC features they would find most and least useful in their everyday creative AI tools.

\begin{table*}[h]
\centering
\caption{Participants' most and least useful PaCC features and their reasoning.}
\label{tab:most_least_useful}
\small
\renewcommand{\arraystretch}{1.4}
\begin{tabularx}{\textwidth}{c l X l X}
\toprule
\textbf{ID} & \textbf{Most Useful} & \textbf{Reason} & \textbf{Least Useful} & \textbf{Reason} \\
\midrule
P1 & Belief Graph & ``It gives me confidence that the complex visual concepts in the scene are exactly how I picture them.'' & --- & ``They are all pretty good.'' \\
P2 & Belief Graph & --- & Attributes & --- \\
P3 & Clar.\ Questions & ``Knowing where the model sees the ambiguity'' & Clar.\ Questions & ``The questions focused on the art style, which is not something I was focused on'' \\
P4 & Clar.\ Questions & --- & Attributes & --- \\
P5 & Clar.\ Questions & ``They elucidate the possible and open things I hadn't been thinking of'' & Belief Graph & ``wasn't particularly useful in my explorations'' \\
P6 & Clar.\ Questions & ``They were relevant and expansive'' & Belief Graph & ``didn't see the need for the belief graph'' \\
P7 & Multi-modal toggle & ``It created a tool to better understand what I'm looking for'' & Belief Graph & ``difficult to interact with and the structure of it didn't reflect how I was thinking'' \\
P8 & Clar.\ Questions \& Attr. & --- & Belief Graph & --- \\
\bottomrule
\end{tabularx}
\end{table*}

\subsection{Additional participant feedback}
\label{sec:appendix_additional}

Selected open-ended responses from participants about their overall experience with PaCC:

\begin{itemize}
    \item \textbf{P1:} ``This was awesome! I would love to see the belief graphs being fed into the network (rather than just text) so I can have more confidence that the complex relationships from the image/video in my head are actually encoded (this would also be nice for image understanding/edit tasks).''
    \item \textbf{P3:} ``This is a great system! Fantastic! I am very excited to see how this develops and how we can better interact with AI to support our agency as humans and artists.''
    \item \textbf{P5:} ``I wish there were a history of my prompts that was more easily accessible. The presence of clarifying questions with multiple choice options makes it desirable to choose something else and see how the output changes.''
    \item \textbf{P6:} ``For a creative tool, the interface was a bit muted, felt like there could be some more delight amidst the UI/UX -- a bit overwhelmed with all the new terminology.''
    \item \textbf{P7:} ``I wonder what is the balance between historic creative methods where there is a feedback loop while you do something visual then look at it. Here I have to imagine what my prompt will create for me visually and I might not know.''
\end{itemize}

\subsection{Effect Sizes}
\begin{table}[t]
\centering
\caption{One-sample Cohen's $d$ ($d = (M-3)/\text{SD}$) for each PaCC feature across agency dimensions, relative to the ``preserved'' midpoint of~3. Because the study had no control condition, these values describe how far ratings lie from the midpoint, not treatment effects of PaCC.}
\label{tab:effect_sizes}
\small
\begin{tabular}{lcccc}
\toprule
Feature & Agency $d$ & Ownership $d$ & Control $d$ & Authorship $d$ \\
\midrule
PaCC (overall)       & $+1.30$ & $+0.15$ & $+1.29$ & $-0.22$ \\
Clarifying Questions & $+1.03$ & $+0.45$ & $+0.83$ & $+0.25$ \\
Belief Graph         & $+0.79$ & $+0.58$ & $+0.54$ & $+0.11$ \\
Attributes           & $+0.35$ & $+0.35$ & $+0.71$ & $+0.07$ \\
Prompt Refinement    & $+0.81$ & $+0.44$ & $+0.54$ & $+0.23$ \\
\bottomrule
\end{tabular}
\end{table}

\subsection{Inter-dimension correlations}
\label{sec:appendix_correlations}

Table~\ref{tab:correlations} reports pairwise Pearson correlations between the four agency dimensions, computed over per-participant mean ratings (aggregated across all features and aspects). Table~\ref{tab:correlations_item} reports the same correlations at the item level (each individual Likert rating).

\begin{table}[h]
\centering
\caption{Pearson correlations between agency dimensions, computed over per-participant mean ratings. Pearson $r$ shows whether ratings rise and fall together across participants, not whether their levels are similar: e.g., Authorship correlates positively with Control even though Authorship ratings are consistently lower.}
\label{tab:correlations}
\small
\begin{tabular}{lcccc}
\toprule
 & Agency & Ownership & Control & Authorship \\
\midrule
\rowcolor{rowlight}
Agency     & --- & $+0.83$ & $+0.95$ & $+0.94$ \\
Ownership  & $+0.83$ & --- & $+0.85$ & $+0.80$ \\
\rowcolor{rowlight}
Control    & $+0.95$ & $+0.85$ & --- & $+0.86$ \\
Authorship & $+0.94$ & $+0.80$ & $+0.86$ & --- \\
\bottomrule
\end{tabular}
\parbox{0.9\textwidth}{\footnotesize \textit{Note.} $N{=}8$ for correlations involving Agency, Ownership, and Control; $N{=}7$ for correlations involving Authorship (one participant did not rate this dimension).}
\end{table}

\begin{table}[h]
\centering
\caption{Pearson ($r$) and Spearman ($\rho$) correlations at the item level (each individual Likert rating paired across dimensions).}
\label{tab:correlations_item}
\small
\begin{tabular}{lcccc}
\toprule
 & $n$ pairs & Pearson $r$ & Spearman $\rho$ \\
\midrule
\rowcolor{rowlight}
Agency $\times$ Ownership   & 152 & $+0.60$ & $+0.61$ \\
Agency $\times$ Control     & 148 & $+0.83$ & $+0.86$ \\
\rowcolor{rowlight}
Agency $\times$ Authorship  & 132 & $+0.63$ & $+0.59$ \\
Ownership $\times$ Control  & 148 & $+0.68$ & $+0.68$ \\
\rowcolor{rowlight}
Ownership $\times$ Authorship & 132 & $+0.69$ & $+0.67$ \\
Control $\times$ Authorship & 128 & $+0.68$ & $+0.61$ \\
\bottomrule
\end{tabular}
\end{table}

\subsection{Extended discussion: Ownership and authorship variation, expressive modality, and role-based hypotheses}
\label{sec:appendix_extended_discussion}

To maintain a concise directional focus in Section~\ref{sec:discussion}, we provide here an extended discussion of three empirical nuances observed in our study: diverging participant perspectives on ownership and authorship, the contrast between conceptual and expressive control, and directional hypotheses regarding user background and role-based needs.

\paragraph{Diverging perspectives on ownership and authorship in generative AI.}
In traditional frameworks of action, sense of agency (SoA), control, ownership, and authorship are closely entwined: feeling in control of an action typically produces a sense of authorship over its outcome~\citep{gallagher2012multiple, moore2016sense}. In our exploratory study, however, participants generally reported high perceived agency ($4.09\pm0.84$) and process control ($3.89\pm0.69$), yet rated authorship lower on average ($2.75\pm1.13$) and expressed polarized views on ownership ($3.22\pm1.49$; Table~\ref{tab:polarization}). Of the six participants who rated both control and authorship, five rated PaCC authorship $0.75$--$1.25$ points below their control rating. Half of our participants (P1, P2, P3, P8) reported high ownership and preserved or augmented authorship (P1 did not rate authorship), whereas the other half (P4--P7) reported low ownership and authorship due to exploring without a pre-formed idea (P5), feeling steered by AI-generated multiple-choice options (P4), skepticism toward AI training data provenance (P6), or text-only style and workflow friction (P5, P7). Notably, individual features also differed: dialogue-based interventions (clarifying questions and prompt refinement) received higher comparative authorship ratings ($5{/}7$ ``greater'') than visualization-based features (belief graph and media attributes). This raises the hypothesis that felt authorship depends on both the creator's initial framing and the \emph{interaction modality} of pre-generation control.

\paragraph{Conceptual versus expressive agency.}
An aspect-level analysis showed that PaCC's pre-generation controls were rated highest for the \emph{conceptual} facets of creative work (intent, process) and lowest for expression, which is \emph{stylistic}. Expression was the lowest-rated aspect for PaCC overall on all four dimensions, including the lowest authorship rating ($2.57$ vs.\ $2.71$--$2.86$). This asymmetry is consistent with the nature of PaCC's interventions: clarifying questions, belief graphs, and prompt refinement all operate in the textual-semantic domain where relational structure and compositional intent can be articulated linguistically. Style is inherently visual and difficult to specify through language alone, as P1 observed in noting that reference images would have helped in confirming which media attributes to update. Text-based controls, while powerful for conceptual agency, may therefore need complementary mechanisms for expressive agency, such as example-based style transfer, to allow users to specify aesthetic intent through visual rather than verbal means. The distinction between conceptual and expressive agency has implications for evaluation, where aggregate agency metrics can mask aspect-level variation, and future evaluations should assess whether AI controls support all facets of creative work or privilege some over others.

\paragraph{Hypotheses on user background and role-based needs.}
Rather than drawing definitive conclusions from our exploratory sample ($N{=}8$), our findings suggest two directional hypotheses regarding individual differences that warrant verification in future work:
\begin{itemize}
    \item \emph{Baseline constraint hypothesis:} Only one participant (P1) reported that current AI tools \emph{very often} diminish their agency (all others answered ``sometimes''), and this participant gave PaCC among the highest agency and control ratings. Though a single case cannot support a correlational claim, it suggests the hypothesis that pre-generation control provides the greatest benefit to creators who feel constrained by reactive post-generation critique loops, whereas users who are comfortable curating AI defaults experience less differentiation. Verifying this hypothesis is important for designing adaptive interfaces that tailor pre-generation scaffolding to individual user workflows.
    \item \emph{Role-based ownership hypothesis:} Averaged across features, AI professionals (P1--P3, P6, P8) and creative professionals (P5, P7) both reported positive agency ($M{=}3.85$ vs.\ $3.72$) and control ($3.80$ vs.\ $3.42$), but the creative professionals rated ownership lower ($2.75$ vs.\ $3.78$), citing the lack of personal visual inputs (P5) or interruptions to their creative flow (P7); skepticism toward AI training data came from an AI professional (P6). This motivates the hypothesis that creative professionals need more direct expressive input and less workflow interruption than technical users before claiming ownership. If validated, designers of creative AI tools should not assume that conceptual steering controls alone confer felt ownership; supporting creative professionals may require transparent provenance, multimodal style inputs from the artist's own work, or hybrid workflows where the AI serves strictly as an ideation partner while the human executes the final artifact.
\end{itemize}

\subsection{Extended discussion of study limitations}
\label{sec:appendix_limitations}
To satisfy reporting requirements while maintaining a directional focus in the main text, we elaborate here on the methodological boundaries of our study. First, our sample size ($N{=}8$) is exploratory and intended to provide directional insights into interactive belief control rather than confirmatory statistical conclusions. Mean agency and control ratings lay well above the scale midpoint (one-sample $d{=}1.30$ and $1.29$), but these describe distance from the midpoint rather than treatment effects relative to a baseline, and individual variation in ownership and authorship underscores the need for larger samples. Second, recruiting primarily AI professionals and technical staff from a single institution means participants possessed high baseline AI literacy; professional artists whose primary livelihood depends on artistic attribution may exhibit different agency thresholds and authorship concerns. Third, the 30-minute task duration focuses on immediate conceptual framing and early ideation; extended, multi-session creative projects could reveal different dynamics around fatigue, trust calibration, and creative ownership. Finally, our comparative evaluation relied on participants' retrospective self-assessment against their familiar tools rather than a controlled A/B experiment against a post-generation critique baseline. We view these limitations as promising roadmaps for future work, including controlled baseline experiments and longitudinal deployments with diverse creative communities.

\newpage

\section{User Study Instructions}
\label{sec:user_study_form}
In this section, we include the full text of instructions given to participants. The user study was done in a Google Form survey.
\subsection*{Section 1: Overview}

Thank you for participating in the PaCC User Study. This survey captures your experience using Proactive Co-Creator (PaCC) and the impact its features have had on your sense of agency. We define Sense of Agency (SAO) in broad terms as \textit{``the feeling of being in control of one's own actions, and their outcomes.''}

\begin{itemize}[noitemsep]
    \item This survey should take 15 minutes.
    \item All questions are optional.
    \item The results of the survey will help us refine PaCC, and will be included in presentations and publications on PaCC.
    \item Results will be aggregated, and any quote attributions will be anonymized.
    \item You may request the removal of your data at any time by emailing the researchers.
\end{itemize}

\vspace{0.5em}
\noindent\textbf{1. Your job role} \\
\underline{\hspace{8cm}}

\vspace{1em}
\noindent\textbf{2. Which of the following would most strongly apply to you? (Select 1)}
\begin{itemize}[label=\radio, noitemsep]
    \item \textbf{Commercial AI artist:} I use AI as my primary tool to generate media in my art practice.
    \item \textbf{Creative Professional:} I am in a creative role (e.g., UX, VisD, Motion Design, etc.) and use AI in my work.
    \item \textbf{AI Professional:} I work in foundational or applied AI Development and Research (e.g., RS, RE, etc.).
    \item \textbf{Other Professional:} I am in a non-creative role (e.g., Sales, Engineering, Legal, etc.) and use AI to supplement my work.
    \item \textbf{Other:} I am in a non-creative role and I don't use AI in my work.
\end{itemize}

\vspace{1em}
\noindent\textbf{3. In the past 6 months (Select all that apply)}
\begin{itemize}[label=\checkbox, noitemsep]
    \item I have used media generating AI to complete a personal art project.
    \item I have used media generating AI to complete a work project.
    \item I have used media generating AI to explore many different ideas for a personal project.
    \item I have used media generating AI to explore many different ideas for a work project.
    \item None of the above.
\end{itemize}

\vspace{1em}
\noindent\textbf{4. In your own words, how would you define ``Sense of Agency''?} \\
\underline{\hspace{12cm}}

\vspace{1em}
\noindent\textbf{5. Consider your creative AI tools or media generation workflows and select the most applicable:}

\vspace{0.5em}
\noindent
\begin{tabularx}{\textwidth}{l *{5}{>{\centering\arraybackslash}X}}
\toprule
 & \textbf{Never} & \textbf{Rarely} & \textbf{Sometimes} & \textbf{Very often} & \textbf{Always} \\
\midrule
The AI tools I use preserve my sense of agency & \radio & \radio & \radio & \radio & \radio \\
The AI tools I use augment my sense of agency & \radio & \radio & \radio & \radio & \radio \\
The AI tools I use diminish my sense of agency & \radio & \radio & \radio & \radio & \radio \\
\bottomrule
\end{tabularx}

\vspace{1em}
\noindent\textbf{6. Provide a brief explanation for your answers above. Include any examples of AI tools you may be thinking of.} \\
\underline{\hspace{12cm}}

\subsection*{Section 2: Your PaCC Experience}
\textit{In this section, we want to understand your overall experience with PaCC.}

\vspace{0.5em}
\noindent\textbf{1. Please consider your experience using PaCC and select the most applicable:}
\begin{itemize}[label=\radio, noitemsep]
    \item PaCC augmented my sense of agency
    \item PaCC increased my sense of agency
    \item PaCC preserved my sense of agency
    \item PaCC reduced my sense of agency
    \item PaCC diminished my sense of agency
\end{itemize}

\vspace{0.5em}
\noindent\textbf{2. Consider your experience using Clarifying Questions in PaCC:}
\begin{itemize}[label=\radio, noitemsep]
    \item Clarifying Questions augmented my sense of agency
    \item Clarifying Questions increased my sense of agency
    \item Clarifying Questions preserved my sense of agency
    \item Clarifying Questions reduced my sense of agency
    \item Clarifying Questions diminished my sense of agency
\end{itemize}

\vspace{0.5em}
\noindent\textbf{3. Consider your experience interacting with the Belief Graph feature in PaCC:}
\begin{itemize}[label=\radio, noitemsep]
    \item The Belief Graph augmented my sense of agency
    \item The Belief Graph increased my sense of agency
    \item The Belief Graph preserved my sense of agency
    \item The Belief Graph reduced my sense of agency
    \item The Belief Graph diminished my sense of agency
\end{itemize}

\vspace{0.5em}
\noindent\textbf{4. Consider your experience using Attributes in PaCC:}
\begin{itemize}[label=\radio, noitemsep]
    \item Attributes augmented my sense of agency
    \item Attributes increased my sense of agency
    \item Attributes preserved my sense of agency
    \item Attributes reduced my sense of agency
    \item Attributes diminished my sense of agency
\end{itemize}

\vspace{0.5em}
\noindent\textbf{5. Consider your experience with Prompt Refinement in PaCC:}
\begin{itemize}[label=\radio, noitemsep]
    \item Prompt refinement augmented my sense of agency
    \item Prompt refinement increased my sense of agency
    \item Prompt refinement preserved my sense of agency
    \item Prompt refinement reduced my sense of agency
    \item Prompt refinement diminished my sense of agency
\end{itemize}

\vspace{0.5em}
\noindent\textbf{6. Provide a brief explanation for your answers above:} \\
\underline{\hspace{12cm}}

\subsection*{Sections 3--7: Feature Assessment \& Creative Alignment}
\textit{Note: The following 1--5 Likert matrices were administered identically across five sections focusing on: \textbf{(Sec. 3) PaCC Overall}, \textbf{(Sec. 4) Clarifying Questions}, \textbf{(Sec. 5) Belief Graph}, \textbf{(Sec. 6) Attributes}, and \textbf{(Sec. 7) Prompt Refinement}.}

\vspace{1em}
\noindent\textbf{1. Impact on your Sense of Agency (1 = Diminished, 3 = Preserved, 5 = Augmented):}
\vspace{0.5em}

\noindent
\begin{tabularx}{\textwidth}{l *{5}{>{\centering\arraybackslash}X}}
\toprule
 & \textbf{1} & \textbf{2} & \textbf{3} & \textbf{4} & \textbf{5} \\
\midrule
Your creative intent, idea, or concept & \radio & \radio & \radio & \radio & \radio \\
Your creative expression and style & \radio & \radio & \radio & \radio & \radio \\
Your creative and exploratory process & \radio & \radio & \radio & \radio & \radio \\
Your final outcome or artifact & \radio & \radio & \radio & \radio & \radio \\
\bottomrule
\end{tabularx}

\vspace{1em}
\noindent\textbf{2. Impact on your Sense of Ownership (1 = Diminished, 3 = Preserved, 5 = Augmented):}
\vspace{0.5em}

\noindent
\begin{tabularx}{\textwidth}{l *{5}{>{\centering\arraybackslash}X}}
\toprule
 & \textbf{1} & \textbf{2} & \textbf{3} & \textbf{4} & \textbf{5} \\
\midrule
Your creative intent, idea, or concept & \radio & \radio & \radio & \radio & \radio \\
Your creative expression and style & \radio & \radio & \radio & \radio & \radio \\
Your creative and exploratory process & \radio & \radio & \radio & \radio & \radio \\
Your final outcome or artifact & \radio & \radio & \radio & \radio & \radio \\
\bottomrule
\end{tabularx}

\vspace{1em}
\noindent\textbf{3. Impact on your Sense of Control (1 = Diminished, 3 = Preserved, 5 = Augmented):}
\vspace{0.5em}

\noindent
\begin{tabularx}{\textwidth}{l *{5}{>{\centering\arraybackslash}X}}
\toprule
 & \textbf{1} & \textbf{2} & \textbf{3} & \textbf{4} & \textbf{5} \\
\midrule
Your creative intent, idea, or concept & \radio & \radio & \radio & \radio & \radio \\
Your creative expression and style & \radio & \radio & \radio & \radio & \radio \\
Your creative and exploratory process & \radio & \radio & \radio & \radio & \radio \\
Your final outcome or artifact & \radio & \radio & \radio & \radio & \radio \\
\bottomrule
\end{tabularx}

\vspace{1em}
\noindent\textbf{4. Impact on your Sense of Authorship (1 = Diminished, 3 = Preserved, 5 = Augmented):}
\vspace{0.5em}

\noindent
\begin{tabularx}{\textwidth}{l *{5}{>{\centering\arraybackslash}X}}
\toprule
 & \textbf{1} & \textbf{2} & \textbf{3} & \textbf{4} & \textbf{5} \\
\midrule
Your creative intent, idea, or concept & \radio & \radio & \radio & \radio & \radio \\
Your creative expression and style & \radio & \radio & \radio & \radio & \radio \\
Your creative and exploratory process & \radio & \radio & \radio & \radio & \radio \\
Your final outcome or artifact & \radio & \radio & \radio & \radio & \radio \\
\bottomrule
\end{tabularx}

\vspace{1em}
\noindent\textbf{5. Provide a brief explanation for your answers above:} \\
\underline{\hspace{12cm}}

\subsection*{Section 8: Comparative Assessment}

\noindent\textbf{1. How would you compare your sense of \underline{agency} while using PaCC to your current AI tools?}
\vspace{0.5em}

\noindent
\begin{tabularx}{\textwidth}{l *{4}{>{\centering\arraybackslash}X}}
\toprule
 & \textbf{Greater than current} & \textbf{Equal to current} & \textbf{Lesser than current} & \textbf{N/A} \\
\midrule
PaCC (overall) & \radio & \radio & \radio & \radio \\
Clarifying Questions & \radio & \radio & \radio & \radio \\
Belief Graph & \radio & \radio & \radio & \radio \\
Attribute Choices & \radio & \radio & \radio & \radio \\
Prompt Refinement & \radio & \radio & \radio & \radio \\
\bottomrule
\end{tabularx}

\vspace{1em}
\noindent\textbf{2. How would you compare your sense of \underline{ownership} while using PaCC to your current AI tools?}
\vspace{0.5em}

\noindent
\begin{tabularx}{\textwidth}{l *{4}{>{\centering\arraybackslash}X}}
\toprule
 & \textbf{Greater than current} & \textbf{Equal to current} & \textbf{Lesser than current} & \textbf{N/A} \\
\midrule
PaCC (overall) & \radio & \radio & \radio & \radio \\
Clarifying Questions & \radio & \radio & \radio & \radio \\
Belief Graph & \radio & \radio & \radio & \radio \\
Attribute Choices & \radio & \radio & \radio & \radio \\
Prompt Refinement & \radio & \radio & \radio & \radio \\
\bottomrule
\end{tabularx}

\vspace{1em}
\noindent\textbf{3. How would you compare your sense of \underline{control} while using PaCC to your current AI tools?}
\vspace{0.5em}

\noindent
\begin{tabularx}{\textwidth}{l *{4}{>{\centering\arraybackslash}X}}
\toprule
 & \textbf{Greater than current} & \textbf{Equal to current} & \textbf{Lesser than current} & \textbf{N/A} \\
\midrule
PaCC (overall) & \radio & \radio & \radio & \radio \\
Clarifying Questions & \radio & \radio & \radio & \radio \\
Belief Graph & \radio & \radio & \radio & \radio \\
Attribute Choices & \radio & \radio & \radio & \radio \\
Prompt Refinement & \radio & \radio & \radio & \radio \\
\bottomrule
\end{tabularx}

\vspace{1em}
\noindent\textbf{4. How would you compare your sense of \underline{authorship} while using PaCC to your current AI tools?}
\vspace{0.5em}

\noindent
\begin{tabularx}{\textwidth}{l *{4}{>{\centering\arraybackslash}X}}
\toprule
 & \textbf{Greater than current} & \textbf{Equal to current} & \textbf{Lesser than current} & \textbf{N/A} \\
\midrule
PaCC (overall) & \radio & \radio & \radio & \radio \\
Clarifying Questions & \radio & \radio & \radio & \radio \\
Belief Graph & \radio & \radio & \radio & \radio \\
Attribute Choices & \radio & \radio & \radio & \radio \\
Prompt Refinement & \radio & \radio & \radio & \radio \\
\bottomrule
\end{tabularx}

\vspace{1.5em}
\noindent\textbf{5. If at all, which PaCC features would you find most useful in your everyday creative AI tools and why?} \\
\underline{\hspace{12cm}}

\vspace{1em}
\noindent\textbf{6. If at all, which PaCC features would you find least useful in your everyday creative AI tools and why?} \\
\underline{\hspace{12cm}}

\vspace{1em}
\noindent\textbf{7. Is there anything else you would like to tell us about your experience today?} \\
\underline{\hspace{12cm}}

\vspace{1em}
\noindent\textbf{8. May we contact you for a follow-up or future studies related to PaCC?}
\begin{itemize}[label=\radio, noitemsep]
    \item Yes
    \item No
\end{itemize}

\newpage
\section*{NeurIPS Paper Checklist}

\begin{enumerate}

\item {\bf Claims}
    \item[] Question: Do the main claims made in the abstract and introduction accurately reflect the paper's contributions and scope?
    \item[] Answer: \answerYes{} %
    \item[] Justification: Claims are justified by references and human studies.
    \item[] Guidelines:
    \begin{itemize}
        \item The answer \answerNA{} means that the abstract and introduction do not include the claims made in the paper.
        \item The abstract and/or introduction should clearly state the claims made, including the contributions made in the paper and important assumptions and limitations. A \answerNo{} or \answerNA{} answer to this question will not be perceived well by the reviewers. 
        \item The claims made should match theoretical and experimental results, and reflect how much the results can be expected to generalize to other settings. 
        \item It is fine to include aspirational goals as motivation as long as it is clear that these goals are not attained by the paper. 
    \end{itemize}

\item {\bf Limitations}
    \item[] Question: Does the paper discuss the limitations of the work performed by the authors?
    \item[] Answer: \answerYes{} %
    \item[] Justification: They are included in discussions.
    \item[] Guidelines:
    \begin{itemize}
        \item The answer \answerNA{} means that the paper has no limitation while the answer \answerNo{} means that the paper has limitations, but those are not discussed in the paper. 
        \item The authors are encouraged to create a separate ``Limitations'' section in their paper.
        \item The paper should point out any strong assumptions and how robust the results are to violations of these assumptions (e.g., independence assumptions, noiseless settings, model well-specification, asymptotic approximations only holding locally). The authors should reflect on how these assumptions might be violated in practice and what the implications would be.
        \item The authors should reflect on the scope of the claims made, e.g., if the approach was only tested on a few datasets or with a few runs. In general, empirical results often depend on implicit assumptions, which should be articulated.
        \item The authors should reflect on the factors that influence the performance of the approach. For example, a facial recognition algorithm may perform poorly when image resolution is low or images are taken in low lighting. Or a speech-to-text system might not be used reliably to provide closed captions for online lectures because it fails to handle technical jargon.
        \item The authors should discuss the computational efficiency of the proposed algorithms and how they scale with dataset size.
        \item If applicable, the authors should discuss possible limitations of their approach to address problems of privacy and fairness.
        \item While the authors might fear that complete honesty about limitations might be used by reviewers as grounds for rejection, a worse outcome might be that reviewers discover limitations that aren't acknowledged in the paper. The authors should use their best judgment and recognize that individual actions in favor of transparency play an important role in developing norms that preserve the integrity of the community. Reviewers will be specifically instructed to not penalize honesty concerning limitations.
    \end{itemize}

\item {\bf Theory assumptions and proofs}
    \item[] Question: For each theoretical result, does the paper provide the full set of assumptions and a complete (and correct) proof?
    \item[] Answer: \answerNA{} %
    \item[] Justification: No theoretical result.
    \item[] Guidelines:
    \begin{itemize}
        \item The answer \answerNA{} means that the paper does not include theoretical results. 
        \item All the theorems, formulas, and proofs in the paper should be numbered and cross-referenced.
        \item All assumptions should be clearly stated or referenced in the statement of any theorems.
        \item The proofs can either appear in the main paper or the supplemental material, but if they appear in the supplemental material, the authors are encouraged to provide a short proof sketch to provide intuition. 
        \item Inversely, any informal proof provided in the core of the paper should be complemented by formal proofs provided in appendix or supplemental material.
        \item Theorems and Lemmas that the proof relies upon should be properly referenced. 
    \end{itemize}

    \item {\bf Experimental result reproducibility}
    \item[] Question: Does the paper fully disclose all the information needed to reproduce the main experimental results of the paper to the extent that it affects the main claims and/or conclusions of the paper (regardless of whether the code and data are provided or not)?
    \item[] Answer: \answerYes{} %
    \item[] Justification: Our human participants are fully anonymized but there is enough information to setup the same human study for other participants.
    \item[] Guidelines:
    \begin{itemize}
        \item The answer \answerNA{} means that the paper does not include experiments.
        \item If the paper includes experiments, a \answerNo{} answer to this question will not be perceived well by the reviewers: Making the paper reproducible is important, regardless of whether the code and data are provided or not.
        \item If the contribution is a dataset and\slash or model, the authors should describe the steps taken to make their results reproducible or verifiable. 
        \item Depending on the contribution, reproducibility can be accomplished in various ways. For example, if the contribution is a novel architecture, describing the architecture fully might suffice, or if the contribution is a specific model and empirical evaluation, it may be necessary to either make it possible for others to replicate the model with the same dataset, or provide access to the model. In general. releasing code and data is often one good way to accomplish this, but reproducibility can also be provided via detailed instructions for how to replicate the results, access to a hosted model (e.g., in the case of a large language model), releasing of a model checkpoint, or other means that are appropriate to the research performed.
        \item While NeurIPS does not require releasing code, the conference does require all submissions to provide some reasonable avenue for reproducibility, which may depend on the nature of the contribution. For example
        \begin{enumerate}
            \item If the contribution is primarily a new algorithm, the paper should make it clear how to reproduce that algorithm.
            \item If the contribution is primarily a new model architecture, the paper should describe the architecture clearly and fully.
            \item If the contribution is a new model (e.g., a large language model), then there should either be a way to access this model for reproducing the results or a way to reproduce the model (e.g., with an open-source dataset or instructions for how to construct the dataset).
            \item We recognize that reproducibility may be tricky in some cases, in which case authors are welcome to describe the particular way they provide for reproducibility. In the case of closed-source models, it may be that access to the model is limited in some way (e.g., to registered users), but it should be possible for other researchers to have some path to reproducing or verifying the results.
        \end{enumerate}
    \end{itemize}

\item {\bf Open access to data and code}
    \item[] Question: Does the paper provide open access to the data and code, with sufficient instructions to faithfully reproduce the main experimental results, as described in supplemental material?
    \item[] Answer: \answerYes{} %
    \item[] Justification: The Proactive Co-Creator is openly accessible via \url{https://proactive-co-creator.github.io/}. The code is front-end only.
    \item[] Guidelines:
    \begin{itemize}
        \item The answer \answerNA{} means that paper does not include experiments requiring code.
        \item Please see the NeurIPS code and data submission guidelines (\url{https://neurips.cc/public/guides/CodeSubmissionPolicy}) for more details.
        \item While we encourage the release of code and data, we understand that this might not be possible, so \answerNo{} is an acceptable answer. Papers cannot be rejected simply for not including code, unless this is central to the contribution (e.g., for a new open-source benchmark).
        \item The instructions should contain the exact command and environment needed to run to reproduce the results. See the NeurIPS code and data submission guidelines (\url{https://neurips.cc/public/guides/CodeSubmissionPolicy}) for more details.
        \item The authors should provide instructions on data access and preparation, including how to access the raw data, preprocessed data, intermediate data, and generated data, etc.
        \item The authors should provide scripts to reproduce all experimental results for the new proposed method and baselines. If only a subset of experiments are reproducible, they should state which ones are omitted from the script and why.
        \item At submission time, to preserve anonymity, the authors should release anonymized versions (if applicable).
        \item Providing as much information as possible in supplemental material (appended to the paper) is recommended, but including URLs to data and code is permitted.
    \end{itemize}

\item {\bf Experimental setting/details}
    \item[] Question: Does the paper specify all the training and test details (e.g., data splits, hyperparameters, how they were chosen, type of optimizer) necessary to understand the results?
    \item[] Answer: \answerNA{} %
    \item[] Justification: Experiments are human studies only.
    \item[] Guidelines:
    \begin{itemize}
        \item The answer \answerNA{} means that the paper does not include experiments.
        \item The experimental setting should be presented in the core of the paper to a level of detail that is necessary to appreciate the results and make sense of them.
        \item The full details can be provided either with the code, in appendix, or as supplemental material.
    \end{itemize}

\item {\bf Experiment statistical significance}
    \item[] Question: Does the paper report error bars suitably and correctly defined or other appropriate information about the statistical significance of the experiments?
    \item[] Answer: \answerYes{} %
    \item[] Justification: We reported standard deviations.
    \item[] Guidelines:
    \begin{itemize}
        \item The answer \answerNA{} means that the paper does not include experiments.
        \item The authors should answer \answerYes{} if the results are accompanied by error bars, confidence intervals, or statistical significance tests, at least for the experiments that support the main claims of the paper.
        \item The factors of variability that the error bars are capturing should be clearly stated (for example, train/test split, initialization, random drawing of some parameter, or overall run with given experimental conditions).
        \item The method for calculating the error bars should be explained (closed form formula, call to a library function, bootstrap, etc.)
        \item The assumptions made should be given (e.g., Normally distributed errors).
        \item It should be clear whether the error bar is the standard deviation or the standard error of the mean.
        \item It is OK to report 1-sigma error bars, but one should state it. The authors should preferably report a 2-sigma error bar than state that they have a 96\% CI, if the hypothesis of Normality of errors is not verified.
        \item For asymmetric distributions, the authors should be careful not to show in tables or figures symmetric error bars that would yield results that are out of range (e.g., negative error rates).
        \item If error bars are reported in tables or plots, the authors should explain in the text how they were calculated and reference the corresponding figures or tables in the text.
    \end{itemize}

\item {\bf Experiments compute resources}
    \item[] Question: For each experiment, does the paper provide sufficient information on the computer resources (type of compute workers, memory, time of execution) needed to reproduce the experiments?
    \item[] Answer: \answerYes{} %
    \item[] Justification: Compute resources are disclosed in \S\ref{ssec:compute} and the exact procedures are described in the human study.
    \item[] Guidelines:
    \begin{itemize}
        \item The answer \answerNA{} means that the paper does not include experiments.
        \item The paper should indicate the type of compute workers CPU or GPU, internal cluster, or cloud provider, including relevant memory and storage.
        \item The paper should provide the amount of compute required for each of the individual experimental runs as well as estimate the total compute. 
        \item The paper should disclose whether the full research project required more compute than the experiments reported in the paper (e.g., preliminary or failed experiments that didn't make it into the paper). 
    \end{itemize}
    
\item {\bf Code of ethics}
    \item[] Question: Does the research conducted in the paper conform, in every respect, with the NeurIPS Code of Ethics \url{https://neurips.cc/public/EthicsGuidelines}?
    \item[] Answer: \answerYes{} %
    \item[] Justification: Internal legal reviews and recommendations were adopted.
    \item[] Guidelines:
    \begin{itemize}
        \item The answer \answerNA{} means that the authors have not reviewed the NeurIPS Code of Ethics.
        \item If the authors answer \answerNo, they should explain the special circumstances that require a deviation from the Code of Ethics.
        \item The authors should make sure to preserve anonymity (e.g., if there is a special consideration due to laws or regulations in their jurisdiction).
    \end{itemize}

\item {\bf Broader impacts}
    \item[] Question: Does the paper discuss both potential positive societal impacts and negative societal impacts of the work performed?
    \item[] Answer: \answerYes{} %
    \item[] Justification: This paper is about the societal impacts related to human agency.
    \item[] Guidelines:
    \begin{itemize}
        \item The answer \answerNA{} means that there is no societal impact of the work performed.
        \item If the authors answer \answerNA{} or \answerNo, they should explain why their work has no societal impact or why the paper does not address societal impact.
        \item Examples of negative societal impacts include potential malicious or unintended uses (e.g., disinformation, generating fake profiles, surveillance), fairness considerations (e.g., deployment of technologies that could make decisions that unfairly impact specific groups), privacy considerations, and security considerations.
        \item The conference expects that many papers will be foundational research and not tied to particular applications, let alone deployments. However, if there is a direct path to any negative applications, the authors should point it out. For example, it is legitimate to point out that an improvement in the quality of generative models could be used to generate Deepfakes for disinformation. On the other hand, it is not needed to point out that a generic algorithm for optimizing neural networks could enable people to train models that generate Deepfakes faster.
        \item The authors should consider possible harms that could arise when the technology is being used as intended and functioning correctly, harms that could arise when the technology is being used as intended but gives incorrect results, and harms following from (intentional or unintentional) misuse of the technology.
        \item If there are negative societal impacts, the authors could also discuss possible mitigation strategies (e.g., gated release of models, providing defenses in addition to attacks, mechanisms for monitoring misuse, mechanisms to monitor how a system learns from feedback over time, improving the efficiency and accessibility of ML).
    \end{itemize}
    
\item {\bf Safeguards}
    \item[] Question: Does the paper describe safeguards that have been put in place for responsible release of data or models that have a high risk for misuse (e.g., pre-trained language models, image generators, or scraped datasets)?
    \item[] Answer: \answerNA{} %
    \item[] Justification: Safeguards are enforced at the model level and inherited by PaCC.
    \item[] Guidelines:
    \begin{itemize}
        \item The answer \answerNA{} means that the paper poses no such risks.
        \item Released models that have a high risk for misuse or dual-use should be released with necessary safeguards to allow for controlled use of the model, for example by requiring that users adhere to usage guidelines or restrictions to access the model or implementing safety filters. 
        \item Datasets that have been scraped from the Internet could pose safety risks. The authors should describe how they avoided releasing unsafe images.
        \item We recognize that providing effective safeguards is challenging, and many papers do not require this, but we encourage authors to take this into account and make a best faith effort.
    \end{itemize}

\item {\bf Licenses for existing assets}
    \item[] Question: Are the creators or original owners of assets (e.g., code, data, models), used in the paper, properly credited and are the license and terms of use explicitly mentioned and properly respected?
    \item[] Answer: \answerYes{} %
    \item[] Justification: The original models are described in \S\ref{ssec:compute}. The license and terms of use follow standard Google AI Studio applets (\url{https://aistudio.google.com/docs/terms}).
    \item[] Guidelines:
    \begin{itemize}
        \item The answer \answerNA{} means that the paper does not use existing assets.
        \item The authors should cite the original paper that produced the code package or dataset.
        \item The authors should state which version of the asset is used and, if possible, include a URL.
        \item The name of the license (e.g., CC-BY 4.0) should be included for each asset.
        \item For scraped data from a particular source (e.g., website), the copyright and terms of service of that source should be provided.
        \item If assets are released, the license, copyright information, and terms of use in the package should be provided. For popular datasets, \url{paperswithcode.com/datasets} has curated licenses for some datasets. Their licensing guide can help determine the license of a dataset.
        \item For existing datasets that are re-packaged, both the original license and the license of the derived asset (if it has changed) should be provided.
        \item If this information is not available online, the authors are encouraged to reach out to the asset's creators.
    \end{itemize}

\item {\bf New assets}
    \item[] Question: Are new assets introduced in the paper well documented and is the documentation provided alongside the assets?
    \item[] Answer: \answerYes{} %
    \item[] Justification: Code released publicly following standard license and terms of use in Google AI Studio.
    \item[] Guidelines:
    \begin{itemize}
        \item The answer \answerNA{} means that the paper does not release new assets.
        \item Researchers should communicate the details of the dataset\slash code\slash model as part of their submissions via structured templates. This includes details about training, license, limitations, etc. 
        \item The paper should discuss whether and how consent was obtained from people whose asset is used.
        \item At submission time, remember to anonymize your assets (if applicable). You can either create an anonymized URL or include an anonymized zip file.
    \end{itemize}

\item {\bf Crowdsourcing and research with human subjects}
    \item[] Question: For crowdsourcing experiments and research with human subjects, does the paper include the full text of instructions given to participants and screenshots, if applicable, as well as details about compensation (if any)? 
    \item[] Answer: \answerYes{} %
    \item[] Justification: See \S\ref{ssec:user_study_design} and \S\ref{sec:user_study_form}.
    \item[] Guidelines:
    \begin{itemize}
        \item The answer \answerNA{} means that the paper does not involve crowdsourcing nor research with human subjects.
        \item Including this information in the supplemental material is fine, but if the main contribution of the paper involves human subjects, then as much detail as possible should be included in the main paper. 
        \item According to the NeurIPS Code of Ethics, workers involved in data collection, curation, or other labor should be paid at least the minimum wage in the country of the data collector. 
    \end{itemize}

\item {\bf Institutional review board (IRB) approvals or equivalent for research with human subjects}
    \item[] Question: Does the paper describe potential risks incurred by study participants, whether such risks were disclosed to the subjects, and whether Institutional Review Board (IRB) approvals (or an equivalent approval/review based on the requirements of your country or institution) were obtained?
    \item[] Answer: \answerYes{} %
    \item[] Justification: The study was exploratory. The work was approved within our institute.
    \item[] Guidelines:
    \begin{itemize}
        \item The answer \answerNA{} means that the paper does not involve crowdsourcing nor research with human subjects.
        \item Depending on the country in which research is conducted, IRB approval (or equivalent) may be required for any human subjects research. If you obtained IRB approval, you should clearly state this in the paper. 
        \item We recognize that the procedures for this may vary significantly between institutions and locations, and we expect authors to adhere to the NeurIPS Code of Ethics and the guidelines for their institution. 
        \item For initial submissions, do not include any information that would break anonymity (if applicable), such as the institution conducting the review.
    \end{itemize}

\item {\bf Declaration of LLM usage}
    \item[] Question: Does the paper describe the usage of LLMs if it is an important, original, or non-standard component of the core methods in this research? Note that if the LLM is used only for writing, editing, or formatting purposes and does \emph{not} impact the core methodology, scientific rigor, or originality of the research, declaration is not required.
    \item[] Answer: \answerYes{} %
    \item[] Justification: As described in the paper, PaCC was built entirely through Google AI Studio applets (\url{https://aistudio.google.com/apps}).
    \item[] Guidelines:
    \begin{itemize}
        \item The answer \answerNA{} means that the core method development in this research does not involve LLMs as any important, original, or non-standard components.
        \item Please refer to our LLM policy in the NeurIPS handbook for what should or should not be described.
    \end{itemize}

\end{enumerate}

\end{document}